# GA-VINO: A Geometry-Aware Variational Physics-Informed Neural Operator for Mindlin–Reissner Plates

**Siqi WANG [a], Daobo SUN [b], Yizheng WANG [c], Yilong ZHANG [b], Yabin JIN [a], Xiaoying ZHUANG [a,d], Timon RABCZUK [a,e*]**

[a] *Institute of Computational Mechanics × AI & College of Intelligent Robotics and Advanced Manufacturing, Fudan University, Shanghai 200433, China*

[b] *School of Aerospace Engineering, Xiamen University, Xiamen, 361102, China*

[c] *Department of Engineering Mechanics, Tsinghua University, Beijing, 100084, China*

[d] *Institute of Photonics, Department of Mathematics and Physics, Leibniz University, Hannover, Germany*

[e] *Institute of Structural Mechanics, Bauhaus-Universität Weimar, Marienstr. 15, D-99423, Weimar, Germany*

**Abstract:** Plate and shell structures are widely used in engineering fields. Rapid response prediction for such structures under complex geometries, heterogeneous materials, and varying loads is important for engineering design, but conventional numerical methods usually require repeated modeling and solution when the physical configuration changes. To address this issue, this study proposes a geometry-aware variational physics-informed neural operator (GA-VINO) for Mindlin–Reissner plates. GA-VINO represents the plate geometry using boundary point clouds and incorporates a material encoder, a load encoder, and a scalar-parameter branch to handle spatially random material fields, spatially varying loads, and sample-level uniform parameters. Through multi-branch point cloud encoding and cross-attention, GA-VINO fuses geometric, material, loading, and query point information, and predicts the transverse deflection and rotations at arbitrary query locations. Unlike conventional data-driven neural operators, GA-VINO requires no labeled solution data during training. Instead, it minimizes a variational physics-informed loss constructed from the discretized total potential energy of the Mindlin–Reissner plate. Compared with grid-based neural operators, GA-VINO directly processes irregular point clouds and allows different physical fields to be discretized on different point sets, avoiding forced interpolation onto a common grid. Numerical results demonstrate that GA-VINO achieves accurate predictions across varying geometries, material properties, and loading conditions, with lower $H^1$ seminorm and energy-norm errors and more accurate predictions of bending moments and transverse shear forces than data-driven baselines. Furthermore, GA-VINO enables millisecond-level full-field inference and demonstrates useful generalization capability under moderate geometric distribution shifts, although its prediction accuracy decreases under stronger geometric extrapolation.
**Keywords:** Mindlin–Reissner plate, Neural operator, Machine learning, Transformer, Partial differential equations

## 1. Introduction

Plate and shell structures are widely used in aerospace [1], marine [2], civil [3], and mechanical engineering systems [4] because of their high stiffness-to-weight ratio and excellent load-carrying efficiency. Accurate prediction of their deformation, rotations, and related mechanical responses is of great importance for structural design, safety assessment, and performance optimization. Many physical phenomena can be modeled by partial differential equations (PDEs) [5][6], and the mechanical response of plate and shell structures is no exception. Numerical methods such as the finite element method (FEM) [7][8], isogeometric analysis (IGA) [9][10], and meshfree methods [11] have been widely used as important tools for solving PDEs and have achieved remarkable success in plate and shell analysis involving complex boundary conditions, material properties, and geometric configurations. However, in many engineering scenarios, structural responses need to be repeatedly evaluated under varying geometries, load distributions, material parameters, or boundary conditions. Such parametric simulations are often essential for design optimization, uncertainty quantification, inverse analysis, and real-time decision making. Although high-fidelity numerical models are reliable, repeatedly solving a large number of plate and shell problems can be computationally expensive, especially when fine meshes, holes, cutouts, or irregular boundaries are involved.

In recent years, AI4PDEs has become a major direction in AI for Science, leading to several deep learning paradigms for solving PDEs. In general, these methods can be categorized into three groups: Physics-informed Neural Networks (PINNs) [12], neural operators [13][14] and Physics-informed Neural Operators (PINOs) [15]. The central idea of PINNs is to use neural networks as universal function approximators to directly solve PDEs, whereas neural operators and PINOs aim to learn mappings between families of input functions and output functions. Several studies have applied strong-form or energy-based PINN methods to predict the responses of plate and shell structures

*Corresponding author

Email addresses: sqwang25@m.fudan.edu.cn (Siqi Wang), timon.rabczuk@uni-weimar.de (Timon Rabczuk)

[16][17][18]. By embedding the governing equations, boundary conditions, or variational energy functionals into the loss function, these methods provide promising alternatives for such problems. However, existing PINN-based methods are essentially designed to solve specific boundary value problems. When the geometry, material type, or boundary condition changes, the network usually needs to be retrained or reoptimized for the new physical configuration. This problem-specific nature limits their efficiency in large-scale parametric studies. Therefore, a more general framework is needed to efficiently predict plate and shell responses under different physical configurations.

Neural operators and PINOs provide promising solutions to this type of problem. According to their architectures, neural operators can be broadly divided into three categories [19]: integral kernel operators based on grids or graphs, deep networks based on universal approximation theory, and Transformer-based neural operators. Among them, the Fourier Neural Operator (FNO) [13] and the Deep Operator Network (DeepONet) [14] are representative methods of the first two categories, respectively. FNO computes kernel integrals using the fast Fourier transform (FFT), which gives it high training efficiency. However, the vanilla FNO is mainly applicable to rectangular domains discretized on uniform grids. Although FNO can be extended to more general geometries through domain embedding, domain mapping, or nearest neighbor strategies, as in DAFNO [20], Geo-FNO [21], and gFNO+ [22], it still essentially performs operator learning on a rectangular computational domain. DeepONet relies on the universal approximation theorem for operators and imposes relatively fewer restrictions on the shape of the computational domain. Nevertheless, its training is usually confined to a specific geometry.

Real-world problems are often characterized by complex and highly variable geometries. To overcome the geometric limitations of the neural operators discussed above, a number of geometry-aware neural operators have been proposed. For example, graph neural operators (GNO) [23] learn input and output functions on graph representations; GINO [24] represents geometry using both signed distance functions (SDFs) and point clouds; Geom-DeepONet [25] incorporates SDFs and Sinusoidal Representation Networks (SIREN) into the DeepONet framework; and Physics-Informed Geometry-Aware Neural Operator (PI-GANO) [26] further integrates physical constraints into geometry-aware operator learning. In addition, several Transformer-based neural operators have been developed for PDE problems on general geometries. The General Neural Operator Transformer (GNOT) [27] encodes PDE parameters and domain geometry as tokens and achieves favorable performance on multiple datasets. Transolver [28] focuses attention on physics-aware tokens, thereby substantially reducing computational complexity. More recently, the Geometry-Informed Neural Operator Transformer (GINOT) [29] represents geometry using boundary point clouds of the computational domain, providing a flexible and scalable approach for solving PDEs on complex geometries.

Although the above methods are capable of solving PDEs on various computational domains, training neural operators usually requires a large amount of data. In many engineering applications, acquiring such data is highly expensive and, in some cases, even impractical. Inspired by PINNs, Li et al. proposed the PINOs. PINOs directly incorporate physical equations into the training process of FNO, substantially reducing the amount of labeled data required for training. In certain cases, they can even be trained without any labeled data. Similarly, PI-DeepONet proposed by Wang et al. [30] can be regarded as a representative physics-informed neural operator based on the DeepONet architecture. However, these methods rely on the strong-form of the governing equations, which inevitably introduces high-order derivatives. Weak form neural operators, such as the finite element operator network [31] and the Variational Physics-informed Neural Operator (VINO) [32], provide an effective way to avoid this issue. VINO directly incorporates the variational form of physical laws into the learning process and can be trained without any labeled data. Nevertheless, since VINO is built on the FNO architecture, its adaptability to different geometries remains limited.

The above studies indicate that geometry-aware neural operators can, to some extent, overcome the limitations of regular grids and fixed computational domains, and PINOs can substantially reduce the dependence on high-fidelity solution labels. However, neural operator architectures that enable label-free training across different computational domains remain very limited. In addition, geometry-aware neural operators for Mindlin–Reissner plates on irregular domains, trained without solution labels, still require further investigation.

To address this issue, we propose a geometry-aware variational physics-informed neural operator (GA-VINO) for Mindlin–Reissner plates. Inspired by GINOT, GA-VINO takes point clouds as input, where the boundary point clouds of the plate mid-surface are used to represent geometric information, thereby greatly improving the flexibility of the model. GA-VINO is trained in an unsupervised physics-driven manner based on the discretized variational functional of Mindlin–Reissner plates, and no labeled data are required during the training process. In addition, a material encoder, a load encoder, and a scalar parameter branch are introduced to provide a unified treatment of different input forms, including nonuniform materials, random loads, and sample-dependent uniform loads. The model directly predicts the transverse deflection and rotations of the Mindlin–Reissner plate at the corresponding query points.

The main contributions of GA-VINO can be summarized as follows:

**Flexible input and output:** GA-VINO decouples the input point clouds from the query points, and employs multiple encoders and a solution decoder to support flexible input of multiple physical sources.

**Purely physics-driven training:** GA-VINO enables unsupervised physics-driven training based on a Transformer architecture and a discretized variational functional, without requiring labeled data.

**Enhanced mechanical consistency:** GA-VINO greatly improves mechanical consistency and reduces gradi-ent-sensitive and energy-based errors.

**Generalization performance:** GA-VINO demonstrates good generalization across different geometries, load distributions, and material properties within the training distribution, and also shows useful predictive capability in geometric out-of-distribution (OOD) problems, particularly under moderate distribution shifts.

To the best of our knowledge, GA-VINO is the first physics-informed neural operator-based solver for thick plates. The remainder of this paper is organized as follows. Section 2 introduces the basic theory of elastic Mindlin–Reissner plates and several paradigms of operator learning. Section 3 presents GINOT and the proposed GA-VINO model. Section 4 provides a comprehensive validation of GA-VINO under different geometries, materials, and load distributions. Finally, Section 5 summarizes the advantages and limitations of GA-VINO and discusses possible directions for future research.

## 2. Preparatory knowledge

In this section, we first introduce the basic theory of elastic Mindlin–Reissner plates. Then, according to whether physical laws are incorporated and how they are incorporated, several paradigms of operator learning are reviewed.

### *2.1. Theory of elastic Mindlin–Reissner plates*

Since the dimension of a plate in the thickness direction is much smaller than its characteristic mid-surface dimensions, appropriate kinematic assumptions are usually introduced based on three-dimensional elasticity theory to reduce the original three-dimensional continuum problem to a two-dimensional mid-surface problem. The classical Kirchhoff–Love thin plate theory takes the plate mid-surface as the reference surface and assumes that straight lines initially normal to the mid-surface remain straight and normal to the deformed mid-surface after deformation. As a result, transverse shear deformation is neglected. This assumption is reasonable for thin plates, but when the plate is relatively thick or transverse shear effects are no longer negligible, the Kirchhoff–Love theory may fail to accurately describe the structural response. The Mindlin–Reissner plate theory extends the Kirchhoff–Love plate theory. Its key assumption is that normals to the mid-surface remain straight after deformation, but they are no longer required to remain perpendicular to the deformed mid-surface. Therefore, the rotations of the plate cross section and the gradients of the mid-surface deflection can be independent of each other, allowing transverse shear strains to exist. This makes the theory suitable for bending analysis of moderately thick plates.

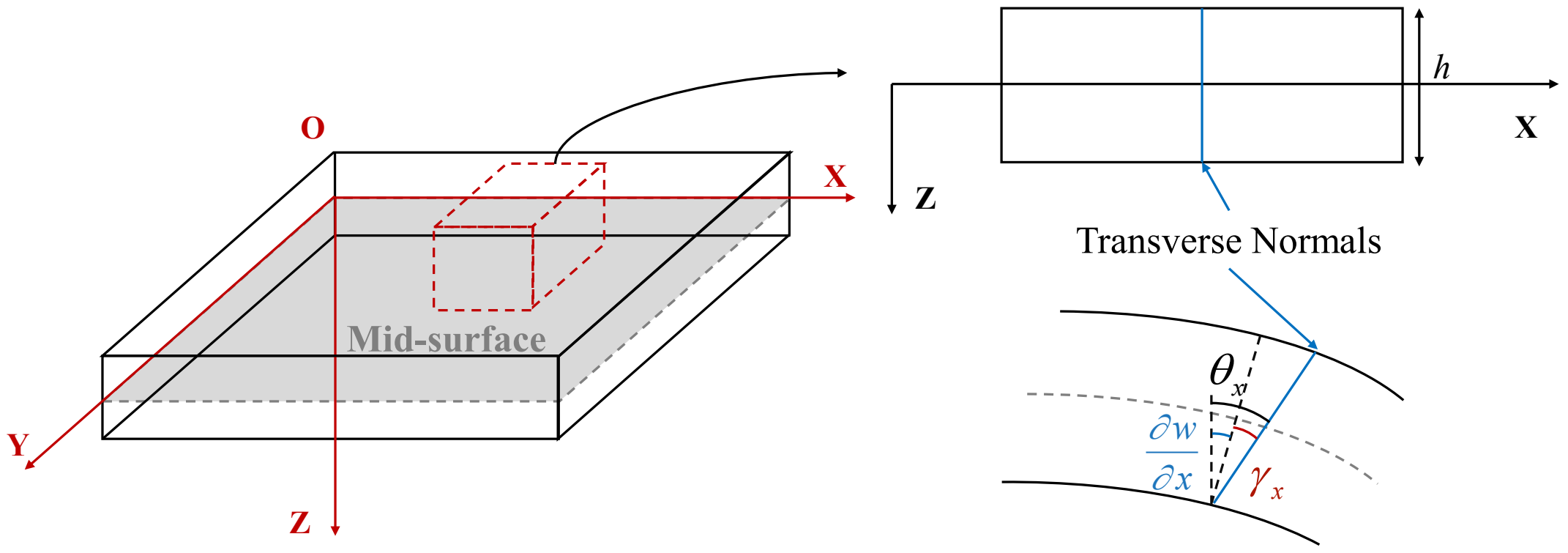

Fig. 1. Kinematic assumptions of the Mindlin–Reissner plate theory. The transverse normals remain straight after deformation but are not required to remain perpendicular to the deformed mid-surface. The difference between the rotation of the deformed mid-surface normal, represented by the deflection gradient $\frac{\partial w}{\partial x}$, and the rotation of the transverse normal $\theta_x$ defines the transverse shear strain component $\gamma_x$.

As shown in Fig. 1, for a Mindlin–Reissner plate undergoing small deformation defined over the mid-surface domain $\Omega \subset R^2$ with thickness $h$, the plate theory uses the transverse deflection $w$ and two rotations $\theta_x$ and $\theta_y$ as independent unknowns to describe the displacement field of the plate. The bending curvature tensor is given by the spatial gradient of the rotation field as

$$\kappa = \nabla^{sym}\theta, \tag{1}$$

where $\theta=[\theta_x, \theta_y]^T$ denotes the rotation vector and $\nabla^{sym}(\cdot)$ denotes the symmetric gradient operator. The transverse shear strain is defined as

$$\gamma = \nabla w - \theta, \tag{2}$$

where $\nabla w$ is the gradient of the transverse deflection. For an isotropic linear elastic material, the Lamé parameters under the plane stress assumption can be expressed as

$$\mu = \frac{E}{2(1+\nu)},$$
$$\lambda = \frac{E\nu}{(1-\nu^2)}, \tag{3}$$

where $E$ is Young's modulus and $\nu$ is Poisson's ratio. The bending strain energy density and the transverse shear strain energy density are respectively given by

$$\psi_{\text{bend}} = \frac{1}{2}\frac{h^3}{12}\left[\lambda\left(\text{tr}\kappa\right)^2 + 2\mu\left(\kappa:\kappa\right)\right], \tag{4}$$

$$\psi_{\text{shear}} = \frac{1}{2}\kappa_s \mu h \gamma \cdot \gamma, \tag{5}$$

where $\kappa_s = 5/6$ is the shear correction factor. The bending moment tensor M and transverse shear force vector Q are obtained from the corresponding strain measures as

$$\text{M} = \frac{h^3}{12}\left[\lambda\left(\text{tr}\kappa\right)\text{I} + 2\mu\kappa\right], \tag{6}$$

$$\text{Q} = \kappa_s \mu h \gamma, \tag{7}$$

where

$$\text{M} = \begin{bmatrix} M_{xx} & M_{xy} \\ M_{xy} & M_{yy} \end{bmatrix}, \tag{8}$$

$$\text{Q} = \begin{bmatrix} Q_x & Q_y \end{bmatrix}^T. \tag{9}$$

The total potential energy of the Mindlin–Reissner plate can be expressed as

$$\Pi = \int_\Omega (\psi_{\text{bend}} + \psi_{\text{shear}}) \text{d}\Omega - \int_\Omega p w \text{d}\Omega - \int_{\Gamma_q} q w \text{d}\Gamma = U_{\text{int}} - W_{\text{ext}} = U_{\text{bend}} + U_{\text{shear}} - W_{\text{ext}}, \tag{10}$$

where $U_{\text{bend}}$ denotes the bending energy, $U_{\text{shear}}$ denotes the transverse shear energy, $p$ is the transverse pressure load, $q$ denotes the transverse edge load applied along the boundary $\Gamma_q$, $w$ is the transverse deflection, $U_{\text{int}}$ is the total strain energy, and $W_{\text{ext}}$ denotes the external work.

*2.2. Neural operators*

The preceding subsection provides the mechanical model and energy functional used in this work. We next review neural operator formulations and physics-informed training paradigms, which form the learning framework of our model.

Neural operators are well suited for mapping infinite-dimensional input functions to output functions evaluated at prescribed query points. This capability has broad applications in science and engineering. For example, the input function may represent the boundary condition of a PDE, while the output corresponds to the solution field of interest. A neural operator can be expressed as

$$\mathcal{G}_\Theta : \mathcal{A} \to \mathcal{U},$$
$$\mathbf{u}_\Theta = \mathcal{G}_\Theta(a), \tag{11}$$

where $\mathcal{A}$ and $\mathcal{U}$ denote the input function space and output function space, respectively, $a \in \mathcal{A}$ denotes the input function, $\mathbf{u}_\Theta \in \mathcal{U}$ denotes the corresponding output function, and $\mathcal{G}_\Theta$ is the neural operator parameterized by the trainable parameters $\Theta$. For a query point $x \in \Omega$, the prediction of the neural operator can be written as

$$\mathbf{u}_\Theta(x) = \mathcal{G}_\Theta(a)(x). \tag{12}$$

In the following, $\mathbf{u}$ denotes the solution field, while $\mathbf{u}_\Theta = \mathcal{G}_\Theta(a)$ denotes its neural-operator approximation. For a given training dataset $\mathcal{D} = \{(a^{(i)}, \mathbf{u}^{(i)})\}_{i=1}^N$, the supervised loss of a purely data-driven neural operator can be written as

$$\mathcal{L}_{\text{data}}(\Theta) = \frac{1}{N_{\text{train}}} \sum_{i=1}^{N_{\text{train}}} \| \mathcal{G}_\Theta(a_i) - \mathbf{u}_i \|_{\mathcal{U}}^2, \tag{13}$$

where $N_{\text{train}}$ is the number of training samples, and $\|\cdot\|$ denotes the $L^2$ norm.

Traditional neural operators are purely data-driven, which makes them simple and flexible. However, training purely data-driven neural operators requires a large amount of high-fidelity data, which significantly limits their engineering applications. In many cases, generating such data may be expensive, unavailable, or restricted to low-resolution observations. To address this difficulty, PINOs have been developed. In the following, we briefly introduce PINOs.

A parametric PDE can be written as

$$\mathcal{N}_a[\mathbf{u}](x) = \boldsymbol{f}(x), \quad x \in \Omega, \tag{14}$$

with the boundary condition

$$\mathcal{B}_a[\mathbf{u}](x) = \boldsymbol{g}(x), \quad x \in \partial\Omega, \tag{15}$$

where $\mathcal{N}_a$ denotes the differential operator parameterized by the input function $a$, $\mathcal{B}_a$ denotes the corresponding boundary operator, $\boldsymbol{f}$ denotes the source term in the domain, and $\boldsymbol{g}$ denotes the prescribed boundary data. The loss function of the strong-form PINOs can be expressed as

$$\begin{aligned} \mathcal{L}_{\text{pde-s}}(\Theta) &= \lambda_{\text{data}}\mathcal{L}_{\text{data}} + \lambda_{\text{pde}}\mathcal{L}_{\text{pde}} + \lambda_{\text{bc}}\mathcal{L}_{\text{bc}}, \\ \mathcal{L}_{\text{pde}} &= \frac{1}{N_r}\sum_{j=1}^{N_r} \| \mathcal{N}_a[\mathbf{u}_\Theta](\mathbf{x}_{r,j}) - \boldsymbol{f}(\mathbf{x}_{r,j}) \|_2^2, \\ \mathcal{L}_{\text{bc}} &= \frac{1}{N_b}\sum_{j=1}^{N_b} \| \mathcal{B}_a[\mathbf{u}_\Theta](\mathbf{x}_{b,j}) - \boldsymbol{g}(\mathbf{x}_{b,j}) \|_2^2, \end{aligned} \tag{16}$$

where $\mathcal{L}_{\text{pde}}$ and $\lambda_{\text{pde}}$, $\mathcal{L}_{\text{data}}$ and $\lambda_{\text{data}}$, and $\mathcal{L}_{\text{bc}}$ and $\lambda_{\text{bc}}$ denote the PDE residual loss, the data loss, the boundary condition loss, and their corresponding weights, respectively. The set $\{x^{(j)}\}_{j=1}^{N_r}$ denotes the residual sampling points in the computational domain, while $\{x^{(j)}\}_{j=1}^{N_b}$ denotes the boundary sampling points. $N_r$ and $N_b$ are the numbers of interior residual points and boundary points, respectively.

In principle, the training of PINOs does not require a data loss. In practice, however, it is often necessary to use labeled data during training, as the strong-form of the governing equations usually involves high-order derivatives [15]. Without the data loss, the accuracy of strong-form PINOs may be insufficient. This motivates the development of VINO [32]. If a parametric PDE admits a variational functional, the corresponding variational loss can be written as

$$\mathcal{L}_{\text{pde-v}}(\Theta) = \frac{1}{N_{\text{train}}}\sum_{i=1}^{N_{\text{train}}}\left[\int_\Omega \mathcal{F}\left(a_i(x), \mathcal{G}_\Theta(a_i)(x), \nabla\mathcal{G}_\Theta(a_i)(x), \cdots\right)\mathrm{d}\Omega + \int_{\delta\Omega} \mathcal{E}\left(a_i(x), \mathcal{G}_\Theta(a_i)(x), \nabla\mathcal{G}_\Theta(a_i)(x), \cdots\right)\mathrm{d}\Gamma\right], \tag{17}$$

where $\mathcal{F}$ and $\mathcal{E}$ denote the variational integral kernels in the interior domain and on the boundary, respectively.

The choice between the strong-form loss $\mathcal{L}_{\text{pde-s}}$ and the variational loss $\mathcal{L}_{\text{pde-v}}$ depends on the nature of the governing PDEs. In general, energy formulations are mainly available for steady-state problems, and not all PDE systems admit a variational principle. Nevertheless, VINO usually achieves better efficiency and accuracy than strong-form PINOs, primarily because variational formulations typically involve lower-order derivatives, which improves the numerical stability of the training process.

## 3. Method

### *3.1. Geometry-Informed Neural Operator Transformer*

The core idea of GINOT is that it does not directly model pairwise self-attention among all input points. Instead, the original geometry point cloud is first encoded into $N_s$ latent tokens through a point cloud geometry encoder, and the query points then interact with these latent tokens through cross-attention. As a result, the computational complexity is reduced from the direct point-to-point interaction form $O(N^2)$ [27] to a form associated with the number of latent tokens, $O(N_s N + N_s^2 + N_q N_s)$. Since $N_s$ is usually much smaller than $N$, GINOT can achieve a significant efficiency advantage. Further discussion on computational complexity is provided in Appendix A.

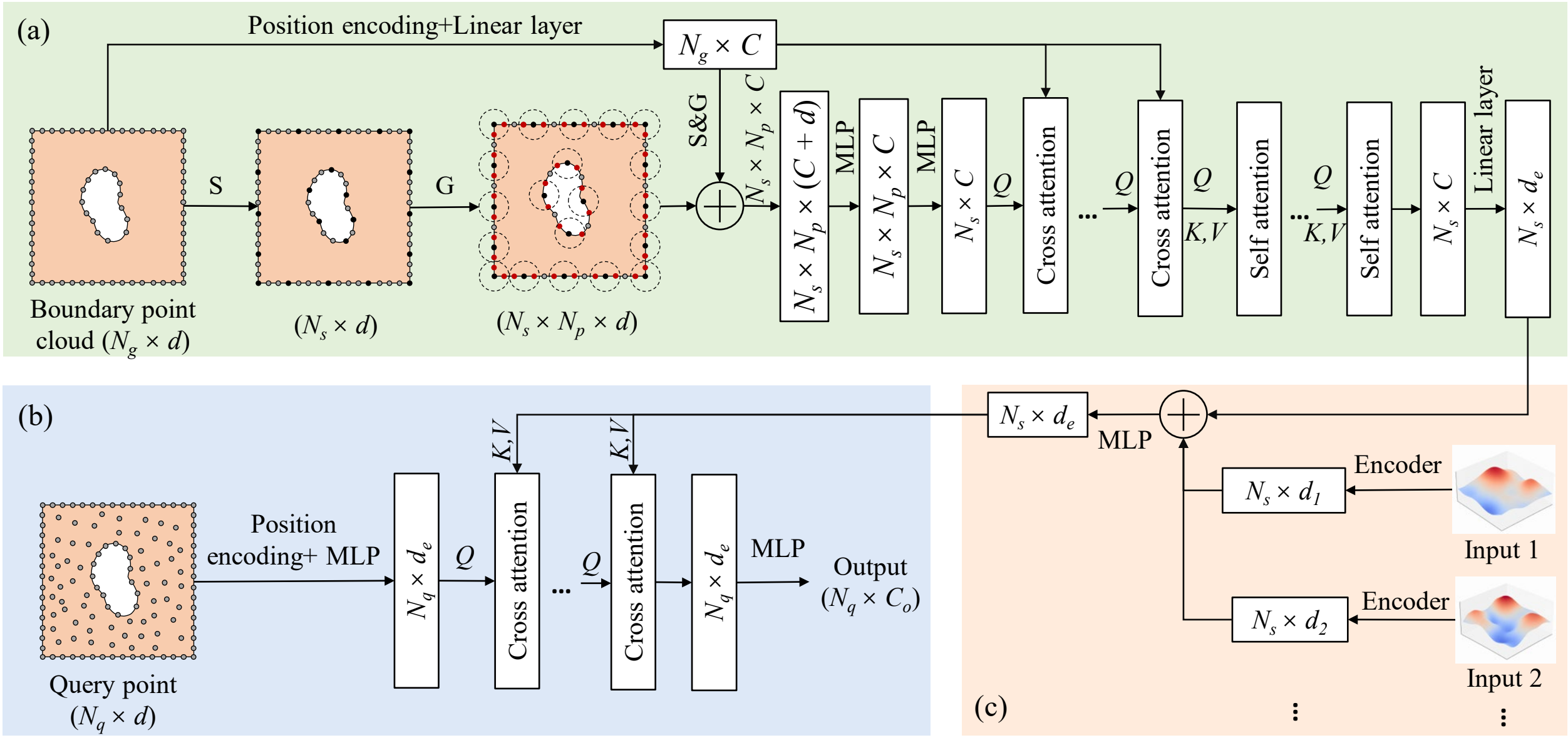

Fig. 2. The architecture of the Geometry-Informed Neural Operator Transformer (GINOT), where S and G denote sampling and grouping, respectively. (a) Geometry encoder based on boundary point clouds. (b) Solution decoder for query point prediction. (c) Additional input encoders for incorporating auxiliary physical fields.

As shown in Fig. 2, GINOT mainly consists of an encoder and a decoder. Both components exploit the attention mechanism inherent in the Transformer architecture, which enables the model to focus on relevant input information. The attention mechanism is defined as

$$\text{Attention}(Q,K,V) = \text{softmax}\left(\frac{QK^T}{\sqrt{d_e}}\right)V, \tag{18}$$

where $Q \in \mathbb{R}^{N_q \times d_e}$ is the QUERY matrix, and $K,V \in \mathbb{R}^{N_k \times d_e}$ are the KEY and VALUE matrices, respectively. Here, $d_e$ denotes the embedding dimension, $N_q$ is the length of the query sequence, and $N_k$ is the length of the KEY and VALUE sequences. Through this mechanism, query points can adaptively attend to the geometric features that are most relevant to their predictions. To improve training stability, residual connections and layer normalization are usually introduced in each attention block.

The geometry encoder takes boundary point clouds as input to extract the shape and boundary topology information of the solution domain. The boundary point cloud is first processed by the sampling layer and the grouping layer. The sampling layer adopts iterative farthest point sampling (FPS). Specifically, one point is first randomly selected from the $N_g$ boundary points, and then the point farthest from the already selected point set is iteratively selected until the number of sampled points reaches $N_s$. Compared with random sampling, this strategy provides better coverage of the geometric boundary. The selected points are then passed to the grouping layer to form $N_s$ point groups, each containing $N_p$ points. For each group, the $N_p$ points are selected within a circle centered at the sampled point with a search radius $r$, which is a hyperparameter. If fewer than $N_p$ points are found within the circle, additional nearest points to the sampled point are included until the number reaches $N_p$. If more than $N_p$ points are found within the circle, the closest $N_p$ points are selected. Therefore, the output size of the grouping layer is $N_s \times N_p \times d$, where $d$ denotes the spatial dimension.

In addition, the boundary point cloud is passed through a Nerf positional encoding layer and a linear layer to obtain higher-dimensional positional features, with an output size of $N_g \times C$, where $C$ denotes the channel dimension after positional encoding. Then, according to the indices obtained from the sampling and grouping operations, the corresponding local geometric information is extracted, resulting in a tensor of size $N_s \times N_p \times C$. This tensor is concatenated with the output of the grouping layer. The combined tensor is then passed through a series of MLP layers to aggregate the information within each group, reducing the $N_p$ dimension to 1. The final output has a size of $N_s \times C$.

In addition to geometry, GINOT can also encode input functions defined on the computational domain. These input functions may represent coefficients, source terms, material parameters, or external load fields. The outputs of these additional encoders are concatenated with the output of the geometry encoder, and an MLP is then used to aggregate information from both the geometry and the additional inputs. The aggregated representation is subsequently used as the KEY and VALUE in the cross-attention blocks of the decoder.

The solution decoder predicts the solution field at query points conditioned on the encoded geometric features and additional inputs. The query points, with a size of $N_q \times d$, are first passed through positional encoding and an MLP to generate a query matrix of size $N_q \times d_e$, where $N_q$ denotes the number of query points and $d_e$ is the embedding dimension. The query matrix is then fed into the cross-attention layers, where geometric information from the geometry encoder is incorporated. The attention mechanism enables the query points to focus on the most relevant geometric features. Finally, another MLP decodes the output of the cross-attention layers into the solution field at the query points.

### *3.2. GA-VINO*

Based on the attention mechanism, GINOT can represent the geometry of complex computational domains using boundary point clouds and predict the corresponding solution function at arbitrary query points. Therefore, it is well suited to irregular geometries. However, GINOT is still essentially a data-driven operator learning framework and usually requires a large number of high-fidelity numerical solutions as supervised data. When such data are expensive to obtain in engineering problems, or when the problem involves a large number of combinations of geometries, materials, and loads, purely data-driven training becomes impractical. In addition, the inputs of Mindlin–Reissner plate problems include not only the geometry, but also material fields, transverse load fields, and sample-dependent mean parameters. Therefore, we propose a geometry-aware variational physics-informed neural operator (GA-VINO) for Mindlin–Reissner plates, whose schematic is shown in Fig. 3.

Compared with conventional neural operators that use grid-based inputs, such as FNO, GA-VINO directly describes geometry, material, and load information in the form of point clouds. For an arbitrary plate domain, the model takes the geometric boundary point cloud, optional material point cloud, optional load point cloud, optional scalar parameters, and a set of query points as inputs, and outputs the plate structural response at the query points. This structure avoids errors associated with interpolating different physical fields onto the same set of points, and reduces the computational cost as much as possible. Its general form can be written as

$$\mathcal{G}_\Theta : (\mathcal{P}_G, \mathcal{P}_M, \mathcal{P}_L, \eta, \mathcal{Q}) \to \mathbf{u}_\Theta(\mathcal{Q}), \tag{19}$$

where $\mathcal{G}_\Theta$ denotes the GA-VINO model with trainable parameters $\Theta$, $\mathcal{P}_G$ denotes the geometric boundary point cloud, $\mathcal{P}_M$ denotes the material point cloud, $\mathcal{P}_L$ denotes the load point cloud, and $\eta$ denotes the sample-level scalar parameter vector. $\mathcal{Q}$ denotes the query point cloud, and $\mathbf{u}_\Theta(\mathcal{Q})$ denotes the Mindlin–Reissner plate response obtained at the query points. Here, $\mathcal{P}_M$, $\mathcal{P}_L$, and $\eta$ are not always present simultaneously, but are selectively activated according to the specific forms of the material and load in the problem considered.

GA-VINO predicts the transverse deflection and two rotational degrees of freedom at each query point. The query point set is denoted by

$$\mathcal{Q} = \{(x^{(q)}, y^{(q)}), q = 1, ..., N_q\}. \tag{20}$$

The output at the $q$-th query point can be expressed as

$$\mathbf{u}_\Theta^{(q)} = [w^{(q)}, \theta_x^{(q)}, \theta_y^{(q)}]^T, \tag{21}$$

where $w^{(q)}$ denotes the transverse deflection at the $q$-th query point, $\theta_x^{(q)}$ and $\theta_y^{(q)}$ denote the corresponding predicted rotations. Therefore, the outputs at all query points can be represented as an $N_q \times 3$ matrix.

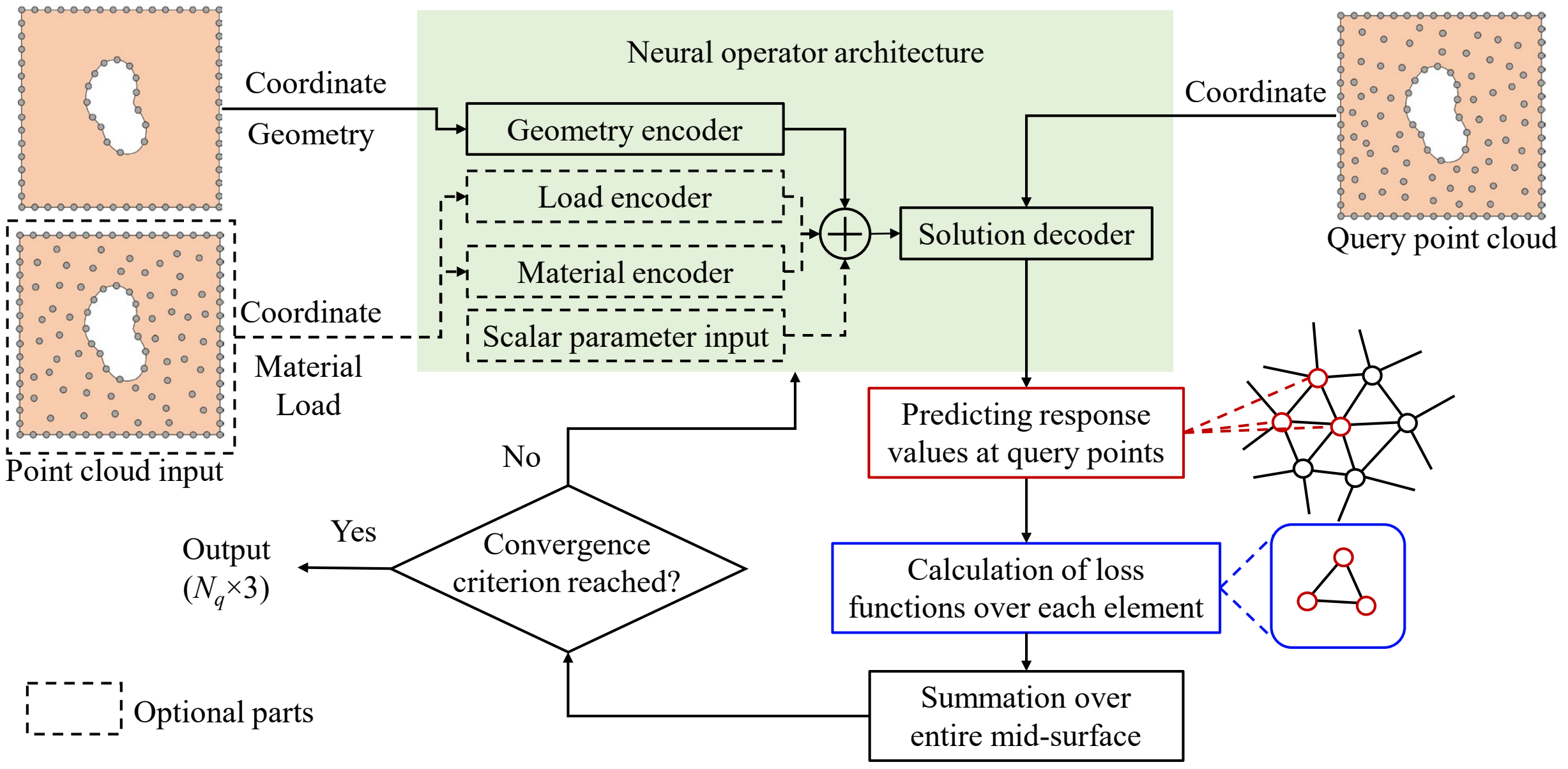

Fig. 3. Schematic of the proposed GA-VINO framework for Mindlin–Reissner plates on irregular domains. The model takes point clouds as input, encodes the plate geometry using boundary point clouds, and selectively incorporates spatially varying material and load fields through dedicated point cloud encoders. Sample-level scalar parameters, such as representative material properties and uniform pressure amplitudes, are processed by a lightweight scalar branch. The fused latent representation is used by the decoder to output the transverse deflection and rotations at arbitrary query points. During training, the predicted nodal responses are evaluated over triangular elements to compute the bending energy, transverse shear energy, and external work. The resulting total potential energy is then minimized as a variational physics-informed loss.

For geometric representation, GA-VINO follows the geometry encoder structure of GINOT and uses the boundary point cloud of the plate mid-surface domain as the geometric input. For a two-dimensional plate domain, the outer boundary, hole boundaries, and cutout boundaries jointly determine the shape and topology of the computational domain. Therefore, the input to the geometry encoder can be expressed as

$$\mathcal{P}_G = \{(x^{(g)}, y^{(g)}), g = 1,\ldots, N_g\}. \tag{22}$$

It should be emphasized that the boundary point cloud and the query point set serve different purposes. The boundary point cloud is used to encode the shape of the computational domain, whereas the query point set specifies the locations where the model outputs the structural response.

GA-VINO can selectively introduce a material encoder to handle heterogeneous material. When the material parameters vary spatially within the domain, the material field is constructed as a point cloud containing both spatial coordinates and physical parameters. The input to the material encoder can be expressed as

$$\mathcal{P}_M = \{(x^{(m)}, y^{(m)}, E^{(m)}, \nu^{(m)}), m = 1,\ldots, N_m\}, \tag{23}$$

where $E^{(m)}$ and $\nu^{(m)}$ denote Young's modulus and Poisson's ratio at the $m$-th material point, respectively, and $N_m$ is the number of points in the material point cloud. For homogeneous materials, if the material parameters are fixed for all samples, the material encoder does not need to be activated. If the material parameters are spatially uniform but their mean values vary among different samples, they can be provided as scalar parameter inputs, without constructing a redundant material point cloud.

Similarly, when the transverse pressure load is spatially nonuniform, GA-VINO can use a load encoder to process the random load. The input to the load encoder can be expressed as

$$\mathcal{P}_L = \{(x^{(l)}, y^{(l)}, P^{(l)}), l = 1,\ldots, N_l\}, \tag{24}$$

where $P^{(l)}$ denotes the transverse pressure value at the $l$-th load point, and $N_l$ is the number of points in the load point cloud. In this case, the load point cloud and the query points may share the same spatial coordinates, but they serve different purposes in the model. The query points are used to specify the locations where the solution is evaluated, whereas the load point cloud is used to encode the input load function. When the load is spatially uniform but its amplitude varies across samples, that is, each sample has a different uniform pressure p, the load is no longer treated as a point cloud input. Instead, it is passed into the scalar MLP branch as a sample-level scalar parameter. Therefore, for materials or loads that are spatially uniform but vary among samples, GA-VINO represents them using a scalar parameter vector

$$\eta = [\overline{E}_{\text{net}}, \overline{\nu}_{\text{net}}, \overline{P}_{\text{net}}], \tag{25}$$

where $\overline{E}_{\text{net}}$, $\overline{\nu}_{\text{net}}$, and $\overline{P}_{\text{net}}$ denote Young's modulus, Poisson's ratio, and the amplitude of the uniform pressure, respectively. In practical applications, if a scalar parameter is fixed for all samples, the corresponding component can either be set as a constant or be deactivated. In this way, GA-VINO can distinguish between spatial random field inputs and sample-level mean parameter inputs. The former are learned by point cloud encoders to capture their spatial distributions, whereas the latter are handled by a lightweight MLP to learn the influence of scalar parameters on the structural response. This avoids redundant point cloud encoding for uniform fields, thereby reducing both the computational cost and GPU memory consumption.

In summary, GA-VINO provides a flexible input and output formulation that can accommodate point cloud inputs with different geometries, different numbers of points, and different spatial distributions. Geometry, material, load, and query points have different physical meanings in the model and can therefore be defined on different point sets according to the specific problem. Meanwhile, when multiple physical quantities are naturally defined at the same spatial locations, the corresponding encoders and decoder can share the same point cloud coordinates. In this way, GA-VINO avoids the restriction of forcing all input fields to be interpolated onto a unified grid or a unified point set, thereby improving its adaptability to complex plate structure problems.

It should be noted that the point-cloud-based neural operator architecture and the discretization required for variational loss evaluation serve different purposes. The geometry encoder and solution decoder operate directly on point clouds without requiring triangular connectivity as a network input. However, during training, a triangular discretization of each computational domain is required to evaluate the total potential energy. The query points serve as the nodes of this discretization, and the corresponding element connectivity is used to compute the bending energy, transverse shear energy, and external work. Different training samples can therefore accommodate different query-point distributions and triangular connectivities. During inference, the trained neural operator directly predicts the response at arbitrary query points without requiring triangular connectivity. Nevertheless, the training discretization can indirectly influence the learned operator through its effect on the discrete variational loss. The sensitivity to the training discretization is further investigated in Section 4.9.

GA-VINO departs from the conventional data-driven training strategy of neural operators and is trained using a variational physics-informed loss. The computational domain is discretized by first-order triangular elements, and shape functions are used to approximate the solution. Derivatives and integrals are then evaluated analytically at the element level. After the model predicts the output $\mathbf{u}_{\Theta}^{(q)}$ at the query points, the bending strain energy and transverse shear strain energy are computed element by element using the triangular element connectivity. Together with the external work, these terms are used to construct the total potential energy functional $\Pi$. For clarity, Appendix B provides the element-wise energy expressions based on three-node linear triangular elements. No labeled data are required during the training of GA-VINO. The network parameters are updated by minimizing the mean total potential energy over all training samples, and the loss function can be expressed as

$$\mathcal{L}(\Theta) = \frac{1}{N_{\text{train}}} \sum_{i=1}^{N} \Pi^{(i)}, \tag{26}$$

where $\Pi^{(i)}$ denotes the total potential energy of the $i$-th training sample, and $N_{\text{train}}$ is the number of training samples. To improve numerical stability, all network input features are normalized before training.

Regarding the treatment of boundary conditions, clamped boundary conditions are imposed through a hard output transformation. GA-VINO first predicts an unconstrained response at each query point $\mathbf{x}_q$:

$$\mathbf{u}_{\Theta}^{\text{raw}}(\mathbf{x}_q) = \left[ w^{\text{raw}}(\mathbf{x}_q), \theta_x^{\text{raw}}(\mathbf{x}_q), \theta_y^{\text{raw}}(\mathbf{x}_q) \right]^T. \tag{27}$$

The final prediction used in the loss function is defined as

$$\mathbf{u}_{\Theta}(\mathbf{x}_q) = \mathbf{m}_{\text{bc}}(\mathbf{x}_q) \odot \mathbf{u}_{\Theta}^{\text{raw}}(\mathbf{x}_q), \tag{28}$$

where $\mathbf{m}_{\text{bc}}(\mathbf{x}_q) \in \{0,1\}^3$ is the boundary-condition mask and $\odot$ denotes component-wise multiplication. For clamped boundary nodes, $\mathbf{m}_{\text{bc}}(\mathbf{x}_q) = [0,0,0]^T$, which exactly enforces $w = 0$, $\theta_x = 0$, and $\theta_y = 0$. For interior nodes and nodes on free boundaries, $\mathbf{m}_{\text{bc}}(\mathbf{x}_q) = [1,1,1]^T$. This transformation is applied before the element-wise energy evaluation; therefore, no additional Dirichlet boundary penalty term is required. For free boundaries, the corresponding natural boundary conditions are incorporated implicitly through the stationarity of the total potential energy, without requiring additional boundary residual terms. When prescribed transverse edge loads are present, their contributions are included in the variational loss through the corresponding boundary work integral.

## 4. Results and discussion

In this section, four groups of numerical experiments are presented to evaluate the performance of GA-VINO. These examples progressively increase in complexity, ranging from homogeneous plates with geometric variations under uniform loading to more general problems involving simultaneous variations in geometry, material properties, and load distributions. These experiments are followed by an evaluation of stress resultants, geometric OOD tests, compari-sons with data-driven and strong-form physics-informed neural operators, and investigations of computational efficiency, training robustness, discretization sensitivity, and thickness effects.

Different input branches are activated according to the physical setting of each problem. For homogeneous plates under fixed uniform loading, only the geometry encoder is used. When the transverse load varies spatially, the load encoder is additionally activated. For heterogeneous plates, the material properties are defined at the centers of the triangular elements formed by neighboring query points and are provided to the model through the material encoder. When a material property or load is spatially uniform but varies in magnitude among samples, it is instead introduced through the scalar-parameter branch.

All experiments are conducted on a single NVIDIA RTX 4090 GPU. To obtain reliable reference solutions and provide a consistent computational comparison, finite element simulations are performed using the open-source FEniCSx-Shells library [33]. The FEM solutions are obtained using a Mindlin–Reissner plate formulation, with GPU acceleration enabled through PETSc and the CUDA-enabled STRUMPACK sparse direct solver. A mesh convergence study is provided in Appendix D to verify the reliability of the FEM reference solutions. The FEM results are used only for accuracy evaluation and for generating the labeled data required by the data-driven baselines. No FEM solution labels are used during the training of GA-VINO.

For Mindlin–Reissner plate problems, evaluating only the relative $L^2$ errors of the transverse deflection and rotations is insufficient to fully assess the mechanical accuracy of the predicted response fields. The bending curvatures depend on the spatial gradients of the rotation field, whereas the transverse shear strains depend on the difference between the deflection gradient and the rotations. Accordingly, six error metrics are adopted to evaluate the accuracy of the predicted kinematic and mechanical fields, including the relative $L^2$ errors of the deflection, rotations, bending moments, and transverse shear forces, the relative $H^1$ seminorm error of the deflection, and the relative energy norm error.

$$\|e_w\|_{L^2} = \frac{\| w^{\text{pred}} - w^{\text{true}} \|_{L^2(\Omega)}}{\| w^{\text{true}} \|_{L^2(\Omega)}}, \tag{29}$$

$$\|e_\theta\|_{L^2} = \frac{\| \theta^{\text{pred}} - \theta^{\text{true}} \|_{L^2(\Omega)}}{\| \theta^{\text{true}} \|_{L^2(\Omega)}}, \tag{30}$$

$$\|e_w\|_{H^1} = \frac{| w^{\text{pred}} - w^{\text{true}} |_{H^1(\Omega)}}{| w^{\text{true}} |_{H^1(\Omega)}}, \tag{31}$$

$$\|e_M\|_{L^2} = \frac{\| \mathrm{M}^{\text{pred}} - \mathrm{M}^{\text{true}} \|_{L^2(\Omega)}}{\| \mathrm{M}^{\text{true}} \|_{L^2(\Omega)}}, \tag{32}$$

$$\|e_Q\|_{L^2} = \frac{\| \mathrm{Q}^{\text{pred}} - \mathrm{Q}^{\text{true}} \|_{L^2(\Omega)}}{\| \mathrm{Q}^{\text{true}} \|_{L^2(\Omega)}}, \tag{33}$$

$$\|e\|_E = \sqrt{\frac{U_{\text{int}}(\mathbf{u}^{\text{pred}} - \mathbf{u}^{\text{true}})}{U_{\text{int}}(\mathbf{u}^{\text{true}})}}. \tag{34}$$

Here, the superscripts "pred" and "true" denote the neural operator predictions and the corresponding FEM reference solutions, respectively. The relative $L^2$ errors of the deflection and rotations measure the accuracy of the primary kinematic fields, while those of the bending moments and transverse shear forces assess the accuracy of the derived mechanical responses. The relative $H^1$ seminorm error evaluates the discrepancy between deflection gradients. The relative energy norm error provides an overall measure of the mechanical discrepancy between the predicted and reference Mindlin–Reissner responses, accounting for both bending and transverse shear contributions.

For the main GA-VINO experiments, the network parameters are optimized using Adam for a maximum of 500 epochs. For the single-hole and double-hole plate examples, early stopping is applied with a minimum of 100 epochs, a patience of 50 epochs, and a relative improvement threshold of 0.1%. The models for the L-shaped and annular-sector plate examples are trained for the full 500 epochs. The main training hyperparameters of GA-VINO are summarized in Table 1, and their effects on prediction performance are further discussed in Appendix C.

Table 2 summarizes the sizes of the geometry, material, load, and query point clouds, together with the per-solution computational costs of GA-VINO inference and the GPU-accelerated FEM solver. As the problem setting becomes more complex, additional encoder branches are activated to process material or load point clouds, resulting in a moderate increase in inference time. Nevertheless, GA-VINO consistently achieves millisecond-level full-field inference, corresponding to online speedups of approximately two to three orders of magnitude over the GPU-accelerated FEM solver. These speedups reflect the online inference stage and do not account for the offline training cost. The corresponding training times and GPU memory consumption are reported in Table 3, while the overall computational efficiency, including the amortization of offline training costs, is further analyzed in Section 4.7.3.

Table 1 Training hyperparameters of the GA-VINO models. UM, RM, UL, and RL denote homogeneous material, heterogeneous material, uniform load, and random load, respectively.

| Hyperparameter | Single-hole plate (UM, UL) | Single-hole plate (UM, RL) | Double-hole plate (UM, UL) | Double-hole plate (UM, RL) | L-shaped plate (RM, UL) | L-shaped plate (RM, RL) | Annular sector (RM, RL) |
|---|---|---|---|---|---|---|---|
| Train/test dataset | 500/100 | 500/100 | 500/100 | 500/100 | 500/100 | 500/100 | 1,000/200 |
| Maximum epochs | 500 | 500 | 500 | 500 | 500 | 500 | 500 |
| Batch size | 16 | 8 | 16 | 8 | 16 | 16 | 16 |
| Initial learning rate | 5e-4 | 5e-4 | 5e-4 | 5e-4 | 5e-4 | 5e-4 | 5e-4 |
| Scheduler patience | 25 | 25 | 25 | 25 | 25 | 25 | 25 |
| Scheduler factor | 0.7 | 0.7 | 0.7 | 0.7 | 0.7 | 0.7 | 0.7 |
| Early stopping | Yes | Yes | Yes | Yes | No | No | No |
| Geometry encoder ($N_s / N_p / r$) | 256/18/0.2 | 384/18/0.2 | 256/18/0.2 | 384/18/0.2 | 256/18/0.2 | 256/18/0.2 | 128/18/0.2 |
| Other encoders ($N_s / N_p / r$) | / | 384/36/0.2 | / | 384/36/0.2 | 256/36/0.2 | 256/36/0.2 | 128/36/0.2 |
| Encoder/Decoder cross-att. layers | 2/4 | 2/4 | 2/4 | 2/4 | 2/4 | 2/4 | 2/4 |
| Encoder self-att. layers | 2 | 2 | 2 | 2 | 2 | 2 | 2 |

Table 2 Input point cloud sizes and computational efficiency of GA-VINO and FEM for different examples. Values in brackets denote the range of point numbers across different samples.

| | Single-hole plate (UM, UL) | Single-hole plate (UM, RL) | Double-hole plate (UM, UL) | Double-hole plate (UM, RL) | L-shaped plate (RM, UL) | L-shaped plate (RM, RL) | Annular sector (RM, RL) |
|---|---|---|---|---|---|---|---|
| $N_g$ | 528 | 528 | [491-592] | [491-592] | 400 | 400 | [198-418] |
| $N_q$ | [10,523-18,821] | [10,523-18,821] | [9,181-17,608] | [9,181-17,608] | [7,393-9,525] | [7,393-9,525] | [2,117-8,874] |
| $N_m$ | / | / | / | / | [14,383-18,648] | [14,383-18,648] | [4,032-17,328] |
| $N_l$ | / | [10,523-18,821] | / | [9,181-17,608] | / | [1,892-2,011] | [4,032-17,328] |
| FEM (s/solution) | 7.06 | 7.03 | 6.27 | 6.23 | 3.86 | 3.89 | 2.43 |
| Inference (s/solution) | 3.61e-3 | 17.63e-3 | 3.16e-3 | 12.35e-3 | 7.34e-3 | 10.89e-3 | 9.70e-3 |

### *4.1. Single-hole plate*

We first consider single-hole plates subjected to transverse pressure loading to evaluate the adaptability of GA-VINO to geometric variations. The plate domain is $\Omega \subset [0,5]^2$, and the thickness is $h = 0.5$ mm. Since the plate thickness is one-tenth of the in-plane dimension, the Kirchhoff–Love plate theory is not suitable for this problem. The Young's modulus is $E = 1000$ MPa and the Poisson's ratio is $\nu = 0.3$. The outer boundary of the plate is fully clamped, and the boundary conditions can be expressed as

$$w = 0, \theta_x = 0, \theta_y = 0, (\text{at x=0, x=5, y=0, y=5}). \tag{35}$$

The size, shape, and location of the hole vary randomly among different samples. The hole center, denoted by $(c_x, c_y)$, can take any position within the region $\Omega_c \subset [1.25, 3.75]^2$. This setting ensures that the entire hole remains inside the square plate and avoids being too close to the outer boundary. The hole boundary is defined by 128 points uniformly distributed in the angular direction. The coordinates of the hole boundary points, denoted by $(x_i, y_i)$, are expressed as

$$\begin{aligned} x_i &= c_x + r_i(\theta_i)\cos(\theta_i), \\ y_i &= c_y + r_i(\theta_i)\sin(\theta_i), \end{aligned} \tag{36}$$

where $r_i$ is the radius of the $i$-th boundary point, and $\theta_i = (2\pi i)/128$. For each sample, the radius $r_i$ can take arbitrary values within a prescribed range, and this range varies from sample to sample. Fig. 4 shows the specific setup of this example, including the geometries and boundary point clouds of different samples.

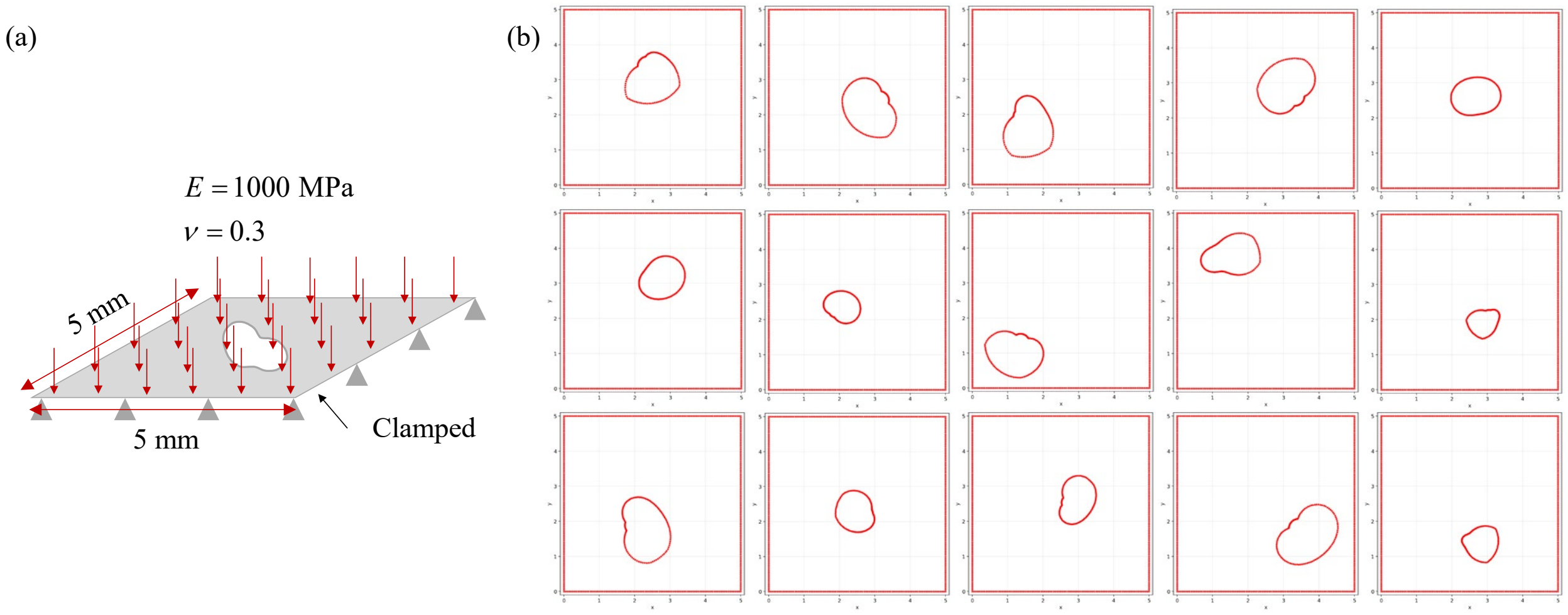

Fig. 4. Single-hole plate under transverse pressure. (a) Loading and boundary condition configuration. (b) Boundary point clouds used as input to the geometry encoder.

### 4.1.1. Homogeneous plate under uniform loading

In this case, the single-hole plate is subjected to a uniform transverse pressure of 0.8 MPa. Since the material parameters and loading amplitude are fixed for all samples, only the geometry encoder is activated. To evaluate the accuracy of the GA-VINO predictions, we compare the deflection $w$ and the rotation magnitude $|\theta| = |\theta_x| + |\theta_y|$ obtained by GA-VINO with the FEM reference solutions.

Fig. 5 (c)–(e) highlight the best case, median case, and worst case in terms of the relative $L^2$ errors of $w$ and $\theta$. For the best case, the relative $L^2$ errors are 0.49% and 2.02%, respectively. For the median case, the relative $L^2$ errors are 1.74% and 3.61%, while the worst case has relative $L^2$ errors of 7.41% and 11.26%. Over all test samples, the mean relative $L^2$ errors are 2.13% and 4.11%, with standard deviations of 1.36% and 2.00%, respectively. The average $H^1$ seminorm error and energy norm error are 4.57% and 11.09%, respectively.

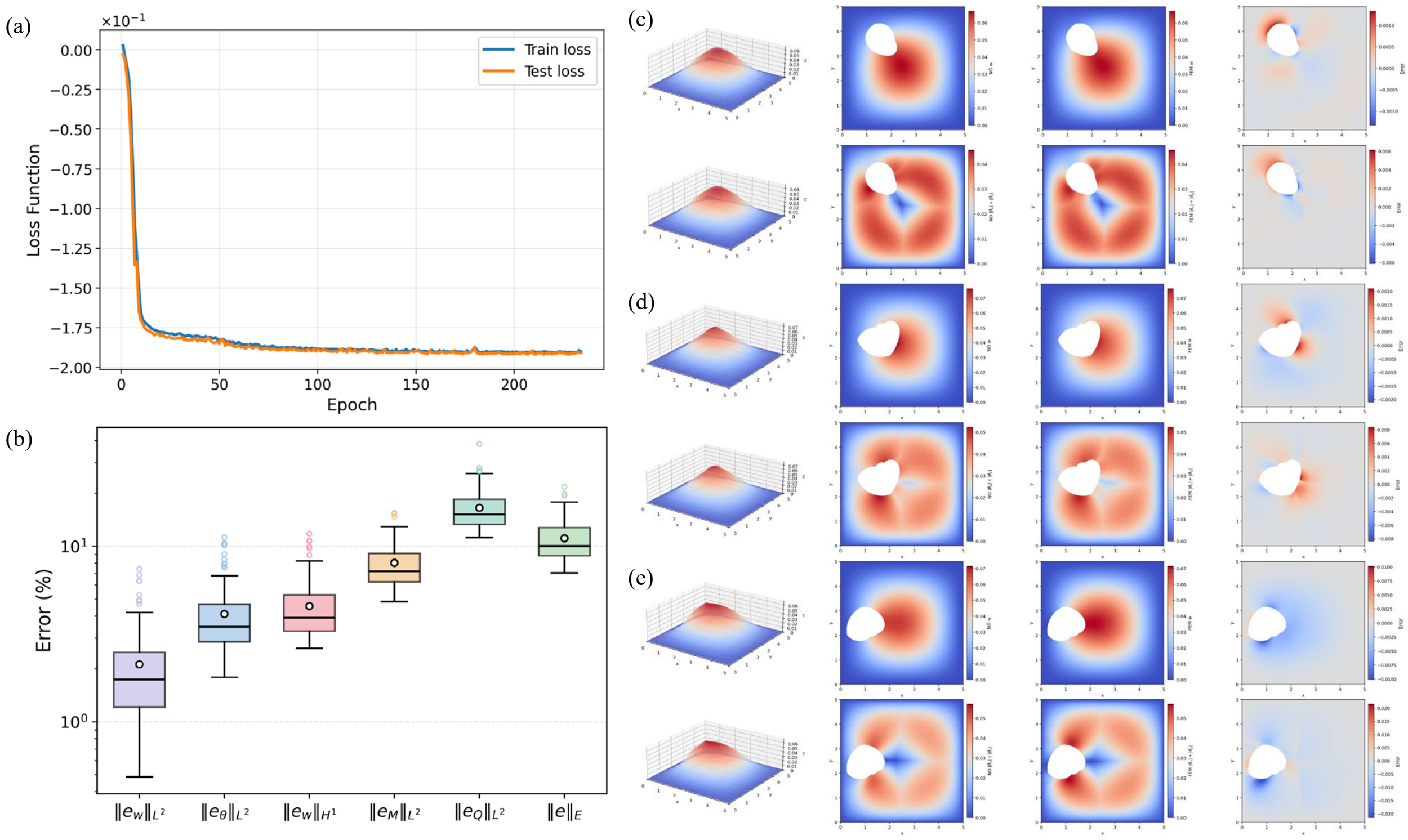

Fig. 5. Single-hole plate with homogeneous material under uniform loading. (a) Evolution of the loss function during training. (b)

Relative errors of the predictions. Distributions of the transverse deflection $w$ and the rotation magnitude $|\theta|$: (c) best case ($\|e_w\|_{L^2} = 0.49\%$, $\|e_\theta\|_{L^2} = 2.02\%$); (d) median case ($\|e_w\|_{L^2} = 1.74\%$, $\|e_\theta\|_{L^2} = 3.61\%$); and (e) worst case ($\|e_w\|_{L^2} = 7.41\%$, $\|e_\theta\|_{L^2} = 11.26\%$). From left to right, the subplots show the surface plot of the transverse deflection, the GA-VINO prediction, the FEM reference solution, and the corresponding error distribution.

### *4.1.2. Homogeneous plate under random load*

For this example, we further evaluate the performance when both the geometry and the load vary among samples. The geometry setting remains the same as that of the single-hole example described above. The load point cloud and the query point cloud share the same spatial coordinates, while the geometry of the Mindlin–Reissner plate is encoded solely by feeding the boundary point cloud into the geometry encoder. The transverse load is sampled from Gaussian random fields (GRFs) and rescaled to the range of 0 MPa to 1.6 MPa, with a mean value of 0.8 MPa. In this case, both the geometry encoder and the load encoder are activated.

Fig. 6 (c)–(e) highlight the best case, median case, and worst case in terms of the relative $L^2$ errors of $w$ and $\theta$. For the best case, the relative $L^2$ errors are 0.64% and 2.15%. For the median case, the relative $L^2$ errors are 1.86% and 3.87%, while the worst case has relative $L^2$ errors of 7.32% and 9.68%. Over all test samples, the mean relative $L^2$ errors are 2.27% and 3.88%, with standard deviations of 1.27% and 1.82%. The average $H^1$ seminorm error and energy norm error are 4.06% and 8.25%, respectively.

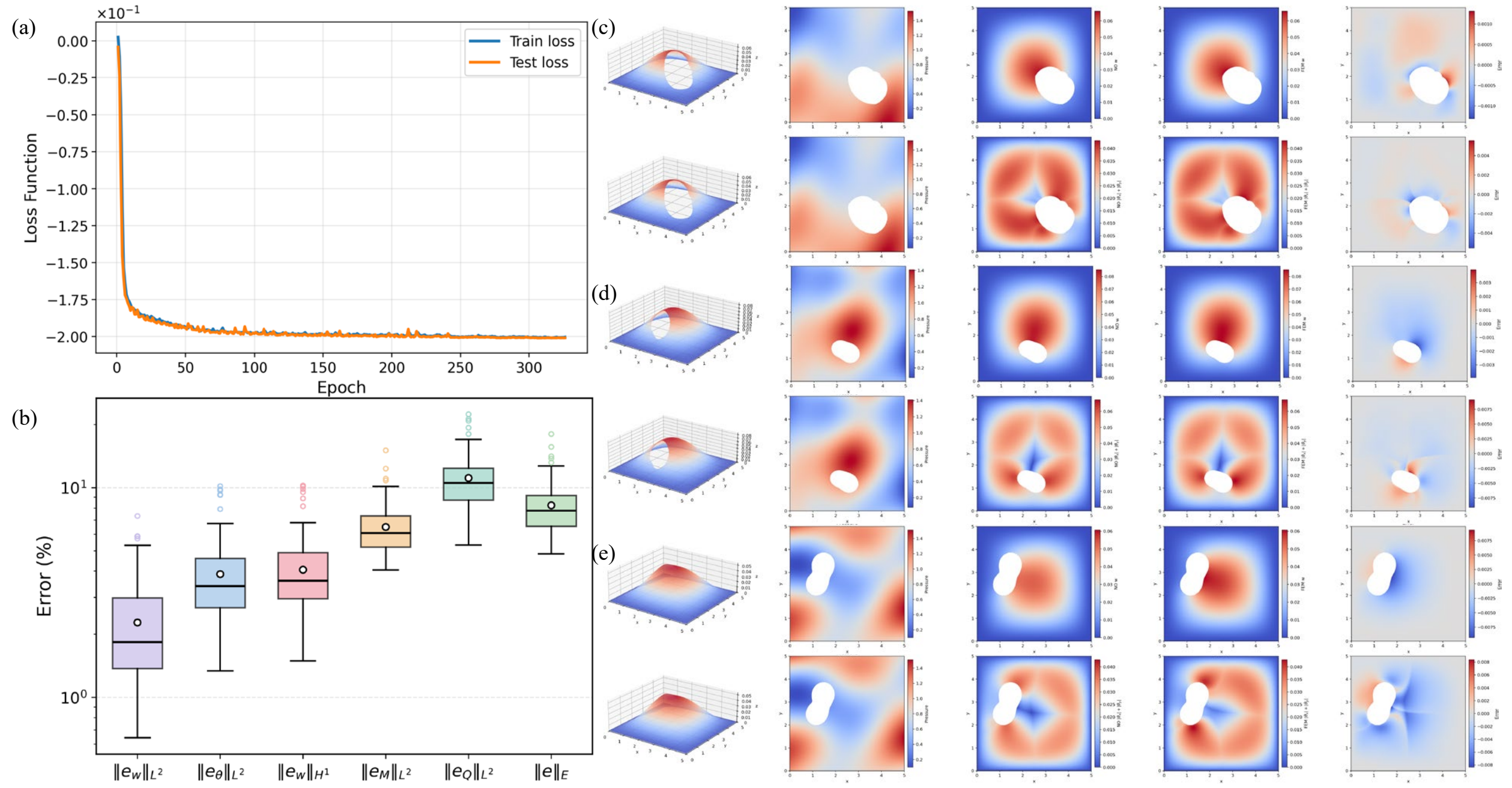

Fig. 6. Single-hole plate with homogeneous material under random load. (a) Evolution of the loss function during training. (b) Relative errors of the predictions. Distributions of the transverse deflection $w$ and the rotation magnitude $|\theta|$: (c) best case ($\|e_w\|_{L^2} = 0.64\%$, $\|e_\theta\|_{L^2} = 2.15\%$); (d) median case ($\|e_w\|_{L^2} = 1.86\%$, $\|e_\theta\|_{L^2} = 3.87\%$); and (e) worst case ($\|e_w\|_{L^2} = 7.32\%$, $\|e_\theta\|_{L^2} = 9.68\%$). From left to right, the subplots show the surface plot of the transverse deflection, the load distribution, the GA-VINO prediction, the FEM reference solution, and the error.

## *4.2. Double-hole plate*

In this section, a double-hole plate example with more pronounced geometric variations is considered to further test the adaptability of GA-VINO to different geometries. The plate domain, thickness, material properties, and clamped boundary conditions are the same as those in the single-hole plate example. Two holes are randomly generated in the plate, with their centers located within the region $\Omega_c \subset [1.25, 3.75]^2$. The distance from each hole boundary point to its corresponding center ranges from 0.2 mm to 1.15 mm. The holes are generated in a manner similar to that used in the single-hole plate example, but here the two holes are allowed to overlap. As a result, the dataset contains richer variations in geometric topology, including separated holes, closely spaced holes, and merged-hole configurations.

Fig. 7 shows the basic setup of this case, including the loading and boundary condition configuration for the uni-form-loading case and representative boundary point clouds used as input to the geometry encoder.

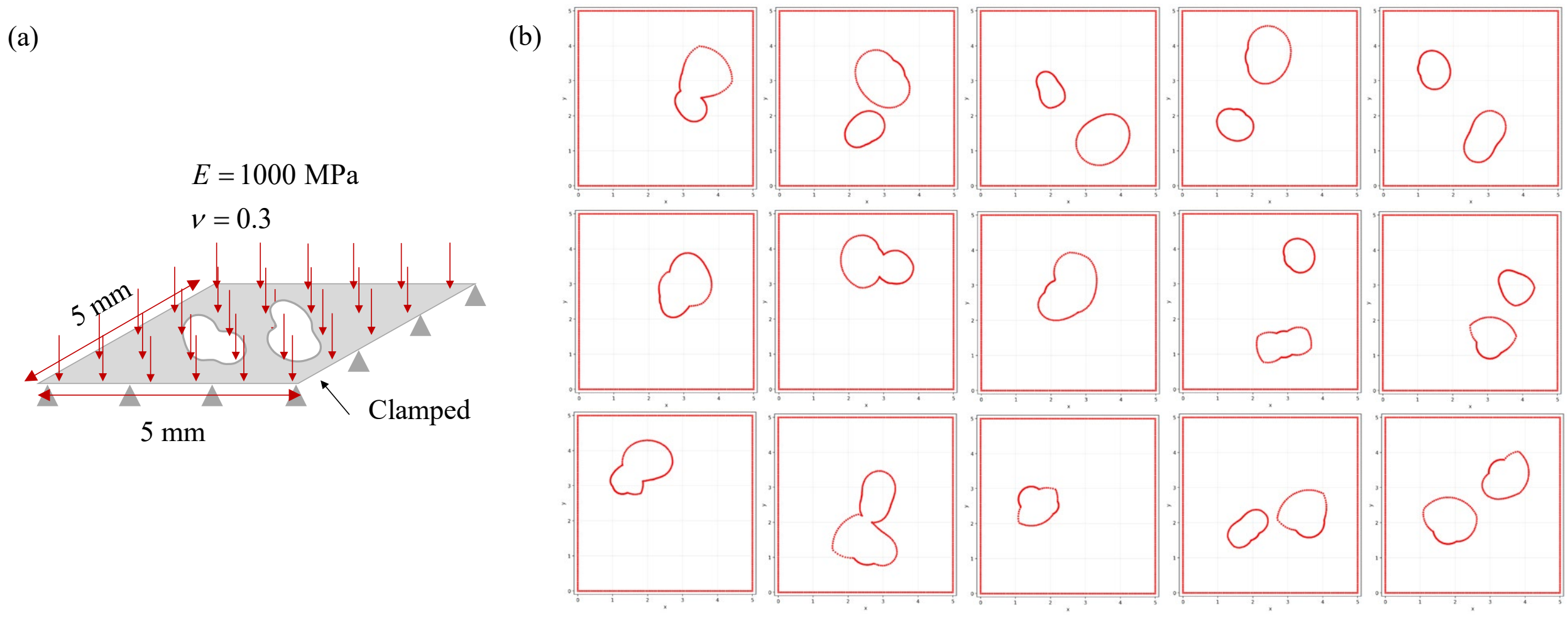

Fig. 7. Double-hole plate under transverse pressure. (a) Loading and boundary condition configuration. (b) Boundary point clouds used as input to the geometry encoder.

### 4.2.1. Homogeneous plate under uniform loading

In this case, the double-hole plate is subjected to a uniform transverse pressure of 0.8 MPa. Since the material parameters and load are fixed for all samples, only the geometry encoder is activated. Fig. 8 (c)–(e) highlight the best case, median case, and worst case in terms of the relative $L^2$ errors of $w$ and $\theta$. For the best case, the relative $L^2$ errors are 0.65% and 2.62%. For the median case, the relative $L^2$ errors are 2.19% and 6.47%, while the worst case has relative $L^2$ errors of 12.48% and 20.07%. Over all test samples, the mean relative $L^2$ errors are 2.94% and 6.28%, with standard deviations of 2.10% and 3.11%. The average $H^1$ seminorm error and energy norm error are 6.51% and 15.39%, respectively.

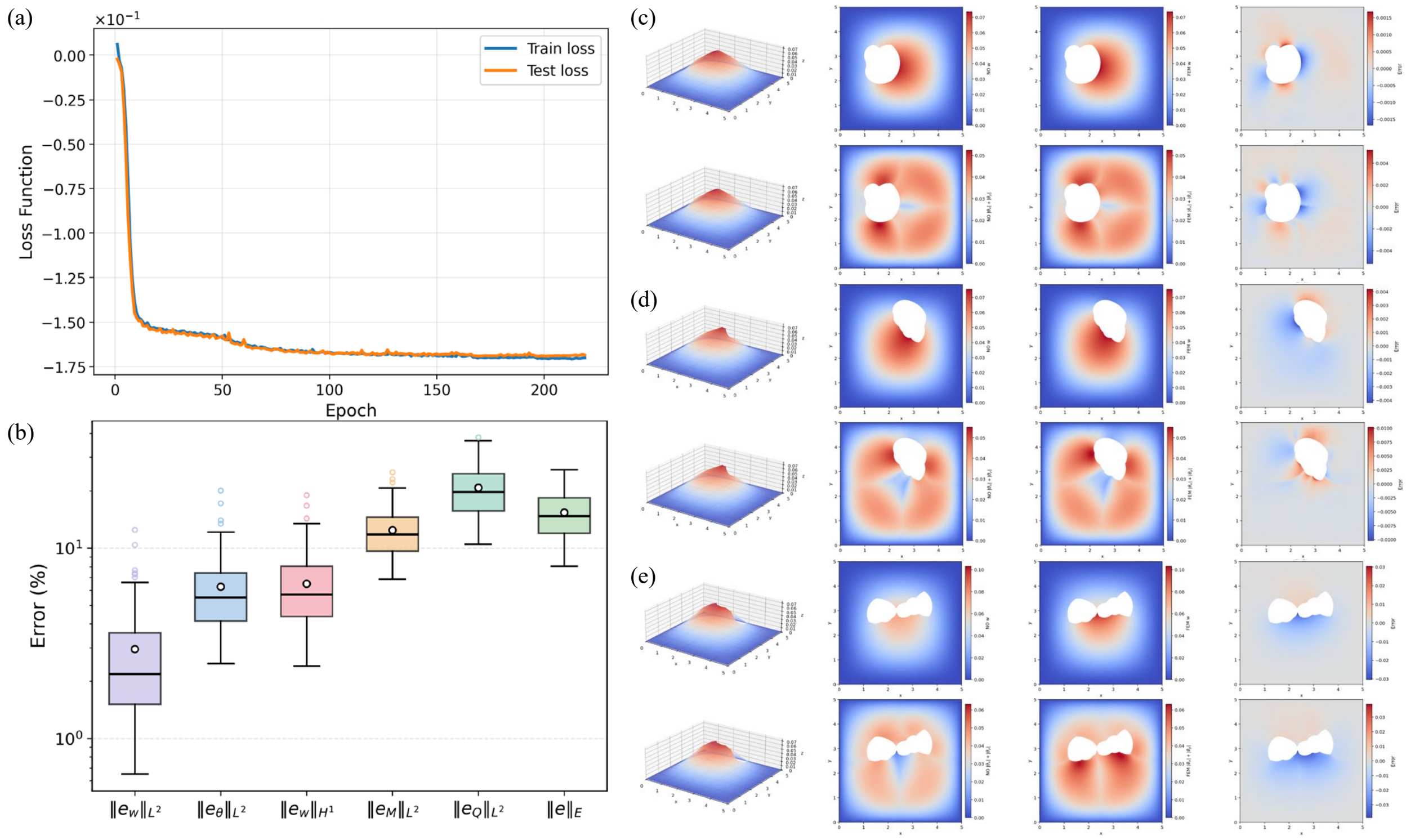

Fig. 8. Double-hole plate with homogeneous material under uniform loading. (a) Evolution of the loss function during training. (b) Relative errors of the predictions. Distributions of the transverse deflection $w$ and the rotation magnitude $|\theta|$: (c) best case

($\|e_w\|_{L^2} = 0.65\%$, $\|e_\theta\|_{L^2} = 2.62\%$); (d) median case ($\|e_w\|_{L^2} = 2.19\%$, $\|e_\theta\|_{L^2} = 6.47\%$); and (e) worst case ($\|e_w\|_{L^2} = 12.48\%$, $\|e_\theta\|_{L^2} = 20.07\%$). From left to right, the subplots show the surface plot of the transverse deflection, the GA-VINO prediction, the FEM reference solution, and the corresponding error distribution.

### *4.2.2. Homogeneous plate under random load*

Similarly, the uniform load is replaced by a spatially varying GRF load, while the double-hole geometry setting remains unchanged. The transverse load is generated from GRFs and rescaled to the range of 0 MPa to 1.6 MPa, with a mean value of 0.8 MPa. In this case, the load encoder is further introduced in addition to the geometry encoder. The transverse deflection and rotation magnitude obtained by GA-VINO and FEM are shown in Fig. 9. For the best case, the relative $L^2$ errors are 1.15% and 3.78%. For the median case, the relative $L^2$ errors are 3.45% and 5.88%, while the worst case has relative $L^2$ errors of 16.90% and 26.14%. Over all test samples, the mean relative $L^2$ errors are 3.98% and 6.97%, with standard deviations of 2.44% and 3.28%. The average $H^1$ seminorm error and energy norm error are 7.10% and 13.69%, respectively. Compared with the single-hole plates, the double-hole cases exhibit larger prediction errors, which may be attributed to stronger local variations in stiffness and deformation induced by closely spaced or overlapping holes.

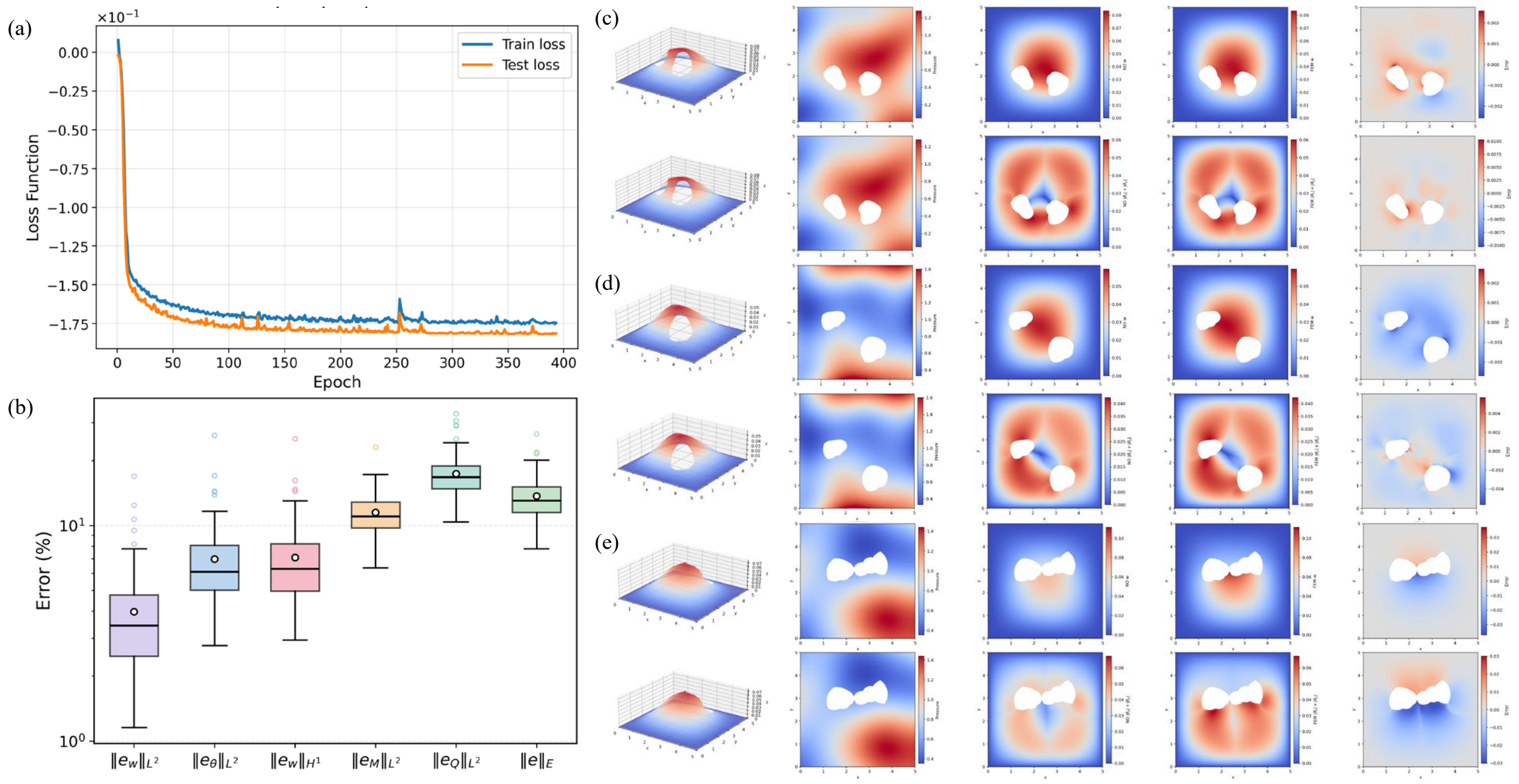

Fig. 9. Double-hole plate with homogeneous material under random load. (a) Evolution of the loss function during training. (b) Relative errors of the predictions. Distributions of the transverse deflection $w$ and the rotation magnitude $|\theta|$: (c) best case ($\|e_w\|_{L^2} = 1.15\%$, $\|e_\theta\|_{L^2} = 3.78\%$); (d) median case ($\|e_w\|_{L^2} = 3.45\%$, $\|e_\theta\|_{L^2} = 5.88\%$); and (e) worst case ($\|e_w\|_{L^2} = 16.90\%$, $\|e_\theta\|_{L^2} = 26.14\%$). From left to right, the subplots show the surface plot of the transverse deflection, the load field distribution, the GA-VINO prediction, the FEM reference solution, and the error.

### *4.3. L-shaped plate*

In this section, we consider an L-shaped plate to further test the predictive capability of GA-VINO when the geometry, material properties, and load vary simultaneously. The L-shaped plate can be regarded as a square plate with a rectangular region of arbitrary size removed from its upper-right corner. The length and width of the removed rectangle are denoted by $l_x$ and $l_y$, with $l_x, l_y \in [1.25, 3.25]$. The left boundary of the plate is clamped, and the boundary condition can be expressed as

$$w = 0, \theta_x = 0, \theta_y = 0, (\text{at x=0}). \tag{37}$$

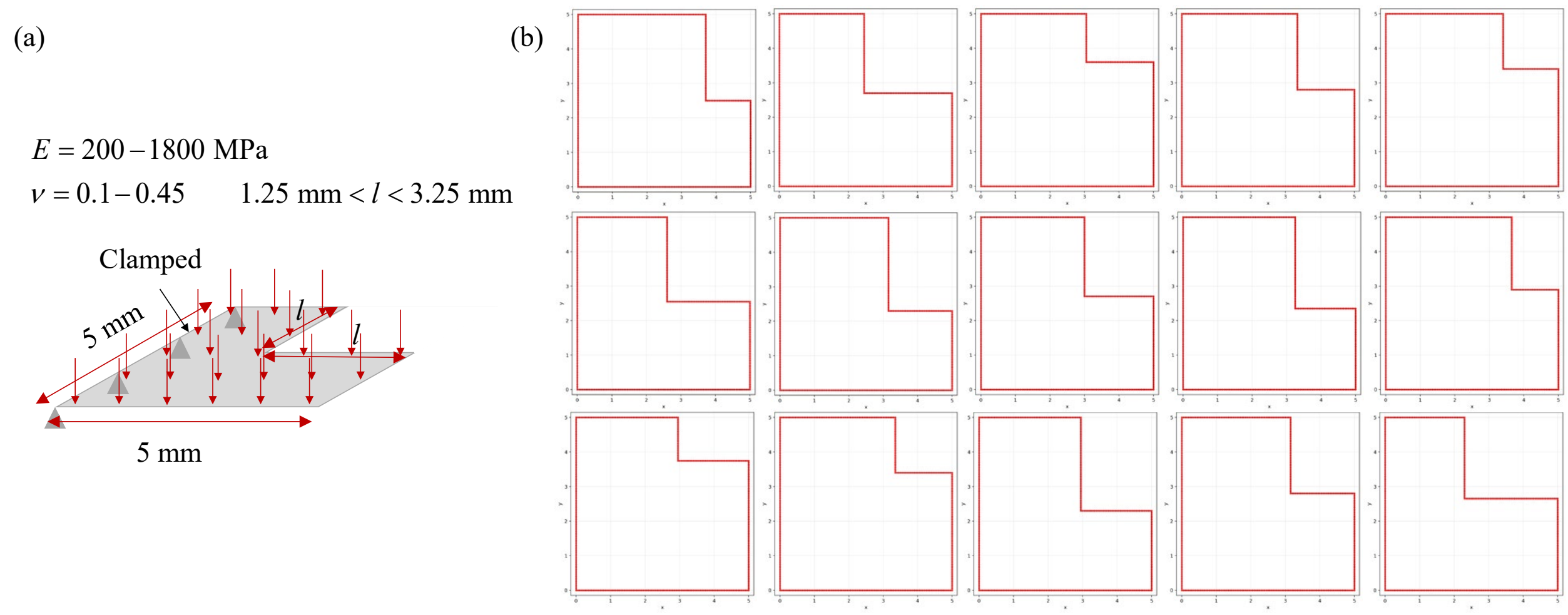

Fig. 10. L-shaped plate under transverse pressure. (a) Loading and boundary condition configuration. (b) Boundary point clouds used as input to the geometry encoder.

The material properties are allowed to vary over a wide range. The distributions of Young's modulus and Poisson's ratio are both generated from GRFs and rescaled to prescribed ranges. Young's modulus varies from 200 MPa to 1800 MPa, and Poisson's ratio varies from 0.1 to 0.45. In this example, the geometry encoder and the material encoder are activated. It should be noted that this example demonstrates a situation in which the encoder input point cloud and the query points are located at different spatial positions. Specifically, the material points are located at the centers of the triangular elements formed by the query points. Fig. 10 shows the basic setup of this case, including the loading and boundary condition configuration for the variable uniform loading case and representative boundary point clouds used as input to the geometry encoder.

### *4.3.1. Heterogeneous plate under variable uniform loading*

In this case, the pressure is spatially uniform, but its amplitude varies randomly from 0.01 MPa to 0.02 MPa among different samples. Therefore, the load is not treated as a point cloud input. Instead, the uniform pressure amplitude is introduced through the scalar parameter branch. The results of this case are shown in Fig. 11. For the best case, the relative $L^2$ errors are 0.25% and 1.15%. For the median case, the relative $L^2$ errors are 1.81% and 2.48%, while the worst case has relative $L^2$ errors of 7.70% and 8.83%. Over all test samples, the mean relative $L^2$ errors are 2.23% and 3.37%, with standard deviations of 1.46% and 1.48%. The average relative $H^1$ seminorm error and energy norm error are 3.47% and 12.17%, respectively.

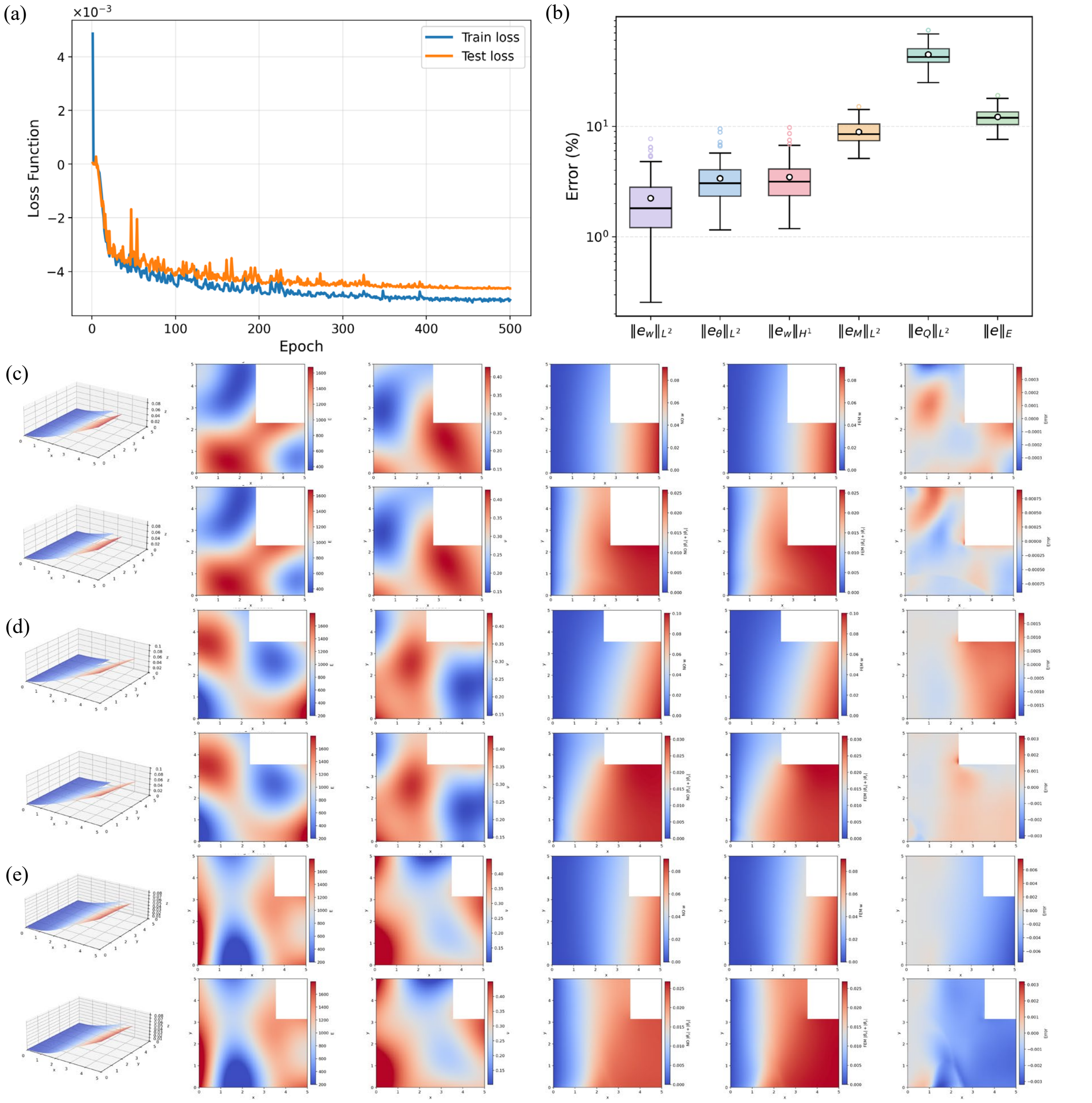

Fig. 11. L-shaped heterogeneous plate under variable uniform loading. (a) Evolution of the loss function during training. (b) Relative errors of the predictions. Distributions of the transverse deflection $w$ and the rotation magnitude $|\theta|$: (c) best case ($\|e_w\|_{L^2} = 0.25\%$, $\|e_\theta\|_{L^2} = 1.15\%$); (d) median case ($\|e_w\|_{L^2} = 1.81\%$, $\|e_\theta\|_{L^2} = 2.48\%$); and (e) worst case ($\|e_w\|_{L^2} = 7.70\%$, $\|e_\theta\|_{L^2} = 8.83\%$). From left to right, the subplots show the surface plot of the transverse deflection, the Young's modulus distribution, the Poisson's ratio distribution, the GA-VINO prediction, the FEM reference solution, and the error.

### *4.3.2. Heterogeneous plate under random load*

We further assume that the load, Young's modulus, and Poisson's ratio all follow GRFs, in order to evaluate the performance of GA-VINO in a more general setting. In this case, the geometry, Young's modulus, Poisson's ratio, and load distribution all vary among samples. The load field is generated from GRFs and provided to the model through the load encoder, while the material fields are still handled by the material encoder. Therefore, the model needs to simultaneously process three types of nonuniform inputs, namely geometry, material properties, and load.

As shown in Fig. 12, the relative $L^2$ errors of the best case are 0.55% and 1.48%. For the median case, the relative $L^2$ errors are 3.04% and 3.19%, while for the worst case, they are 14.44% and 14.75%. Over all test samples, the mean relative $L^2$ errors are 3.61% and 4.96%, with standard deviations of 2.47% and 2.35%. The average $H^1$ seminorm error and energy norm error are 5.00% and 13.28%, respectively.

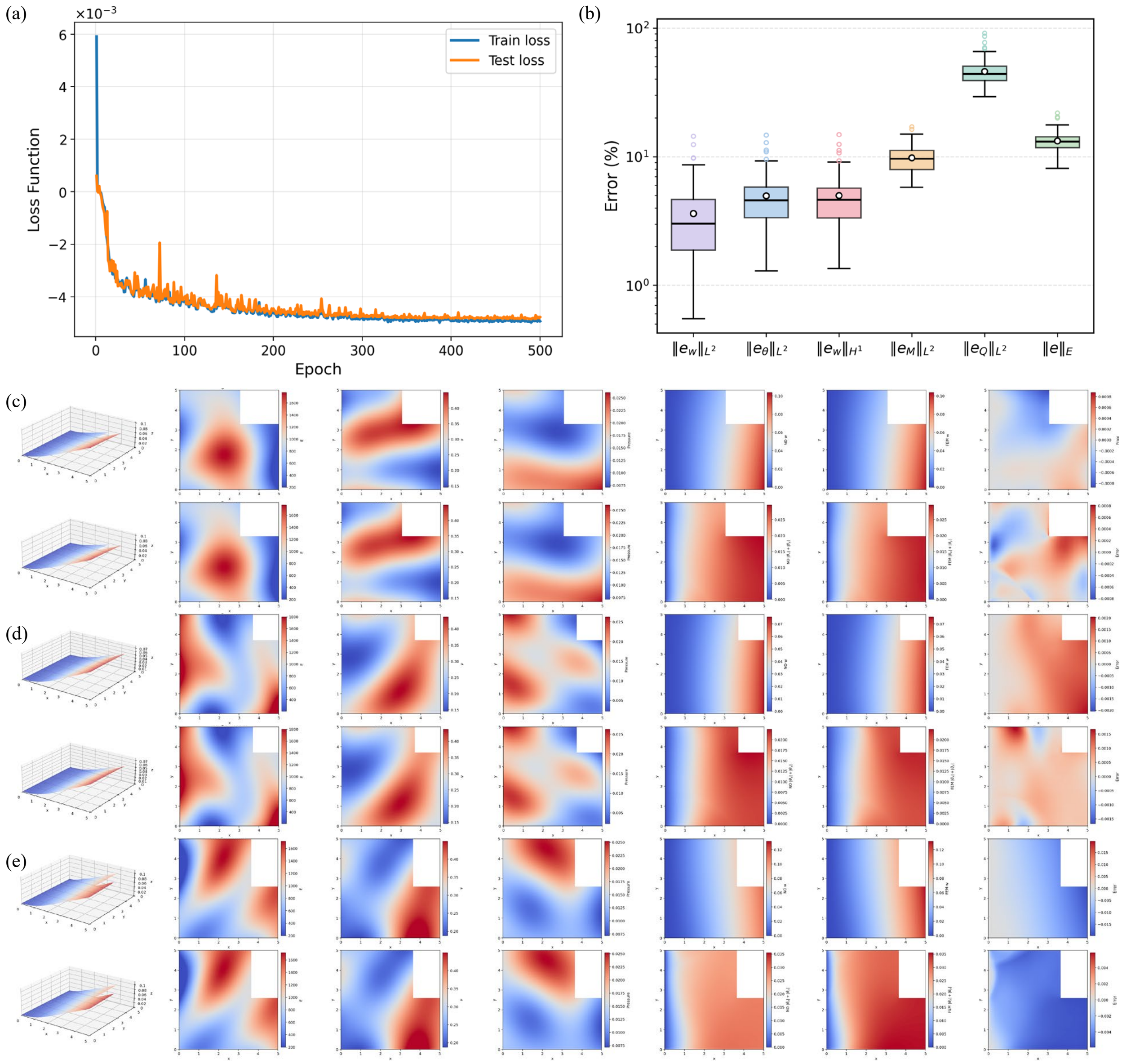

Fig. 12. L-shaped heterogeneous plate under random load. (a) Evolution of the loss function during training. (b) Relative errors of the predictions. Distributions of the transverse deflection $w$ and the rotation magnitude $|\theta|$: (c) best case ($\|e_w\|_{L^2} = 0.55\%$, $\|e_\theta\|_{L^2} = 1.48\%$); (d) median case ($\|e_w\|_{L^2} = 3.04\%$, $\|e_\theta\|_{L^2} = 3.19\%$); and (e) worst case ($\|e_w\|_{L^2} = 14.44\%$, $\|e_\theta\|_{L^2} = 14.75\%$). From left to right, the subplots show the surface plot of the transverse deflection, the Young's modulus distribution, the Poisson's ratio distribution, the load field distribution, the GA-VINO prediction, the FEM reference solution, and the error.

### *4.4. Annular sector plate*

Finally, we consider annular sector plates with varying geometries subjected to spatially varying transverse edge loads to further evaluate the capability of GA-VINO in handling curved boundaries, heterogeneous materials, and random loads defined along the boundary. The centers of the annular sector plates are located at the origin. The inner radius $r_i$ is randomly sampled from 2.0 mm to 3.0 mm, while the outer radius $r_o$ is randomly sampled from 4.0 mm to 5.0 mm. The end angle is fixed at $\theta_e$, and the starting angle $\theta_s$ is randomly sampled from 0 to $\pi/4$, corresponding to an angular span ranging from $\pi/4$ to $\pi/2$. Therefore, different samples have different radial dimensions and angular spans. The computational domain can be expressed as

$$\Omega = \{(r\cos\theta, r\sin\theta) \,|\, r_i \le r \le r_o, \theta_s \le \theta \le \theta_e\}, \tag{38}$$

where $\theta_e = \pi/2$ and $0 < \theta_s < \pi/4$. The plate thickness is set to $h = 0.5$ mm. The left boundary of the plate is clamped, and the boundary conditions can be expressed as

$$w = 0, \theta_x = 0, \theta_y = 0, (\text{at x=0}). \tag{39}$$

In this example, Young's modulus and Poisson's ratio are both generated from GRFs and rescaled to prescribed ranges. Young's modulus varies from 200 MPa to 1800 MPa, while Poisson's ratio varies from 0.10 to 0.45. The boundary at $\theta = \theta_s$ is subjected to a spatially varying transverse edge load $q$, whose values are restricted to the range of 0 to 0.06 N/mm, with a mean value of 0.02 N/mm. Unlike the preceding examples involving distributed pressure loads, the external work in this case is evaluated through the boundary integral in Eq. (10), with the domain pressure set to zero. The corresponding discrete evaluation is detailed in Appendix B.

To represent the transverse edge load in the load encoder, the load is embedded into a point-cloud representation defined over the entire computational domain. Nodes located on the loaded boundary are assigned their corresponding line-load values, whereas the remaining nodes are assigned zero. The zero-valued entries are introduced only to preserve the spatial coverage required by the point-cloud sampling and to enable the load encoder to identify the spatial location and distribution of the applied boundary load. Considering the increased complexity of this problem, 1,000 training samples and 200 test samples are used for model training and testing, respectively. Fig. 13 shows the specific setup of this example, including representative geometries and boundary point clouds for different samples.

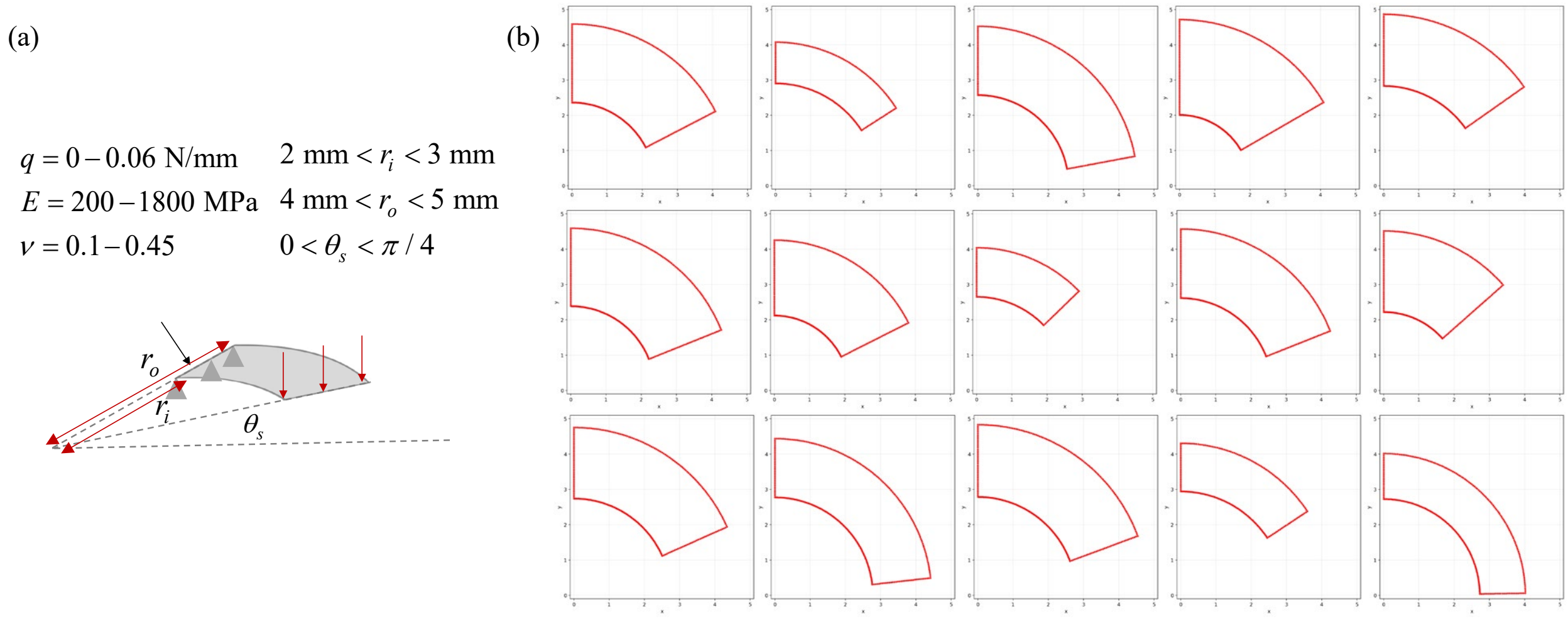

Fig. 13. Annular sector plate under transverse edge loads. (a) Loading and boundary condition configuration. (b) Boundary point clouds used as input to the geometry encoder.

Fig. 14 shows the results of this example. For the best case, the relative $L^2$ errors of the deflection and rotation fields are 0.52% and 2.35%, respectively. For the median case, the corresponding errors are 3.21% and 3.92%, while the worst case has relative $L^2$ errors of 13.51% and 11.93%. Over all test samples, the mean relative errors are 3.93% and 4.62%, with standard deviations of 2.64% and 2.28%, respectively. The average relative $H^1$ seminorm error and energy norm error are 5.15% and 16.21%, respectively.

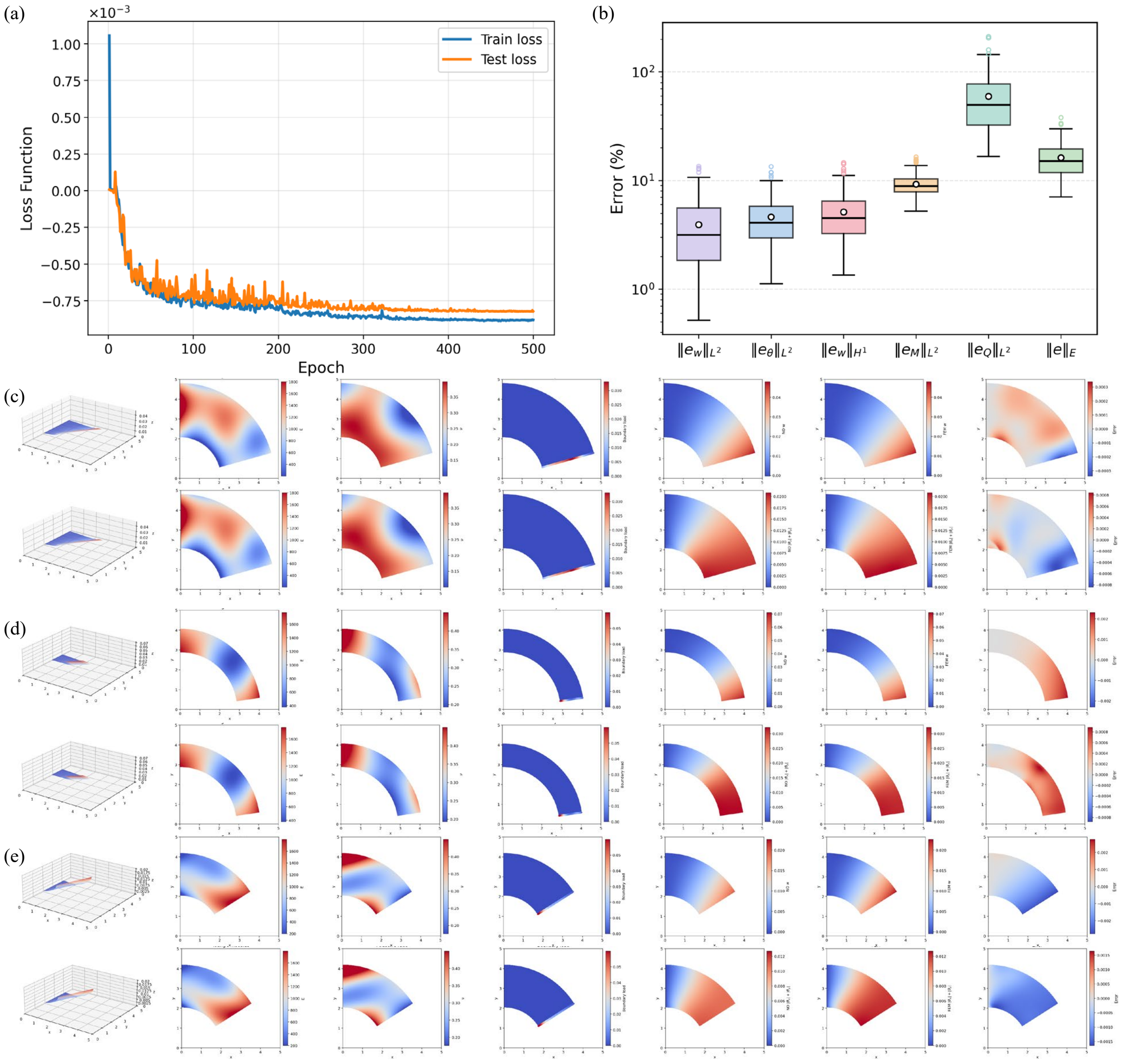

Fig. 14. Annular sector heterogeneous plate under transverse edge loads. (a) Evolution of the loss function during training. (b) Relative errors of the predictions. Distributions of the transverse deflection $w$ and the rotation magnitude $|\theta|$: (c) best case ($\|e_w\|_{L^2} = 0.52\%$, $\|e_\theta\|_{L^2} = 2.35\%$); (d) median case ($\|e_w\|_{L^2} = 3.21\%$, $\|e_\theta\|_{L^2} = 3.92\%$); and (e) worst case ($\|e_w\|_{L^2} = 13.51\%$, $\|e_\theta\|_{L^2} = 11.93\%$). From left to right, the subplots show the surface plot of the transverse deflection, the Young's modulus distribution, the Poisson's ratio distribution, the transverse edge load distribution, the GA-VINO prediction, the FEM reference solution, and the error.

### *4.5. Evaluation of stress resultants*

In addition to transverse deflection and rotations, bending moments and transverse shear forces are important mechanical quantities in plate analysis. According to Eqs. (6)–(9), these stress resultants are determined by the primary kinematic variables and their spatial derivatives. Therefore, small prediction errors in deflection and rotations do not necessarily imply comparable accuracy in bending moments and shear forces. To further evaluate the capability of GA-VINO to predict these derived mechanical responses, the homogeneous single-hole plate and the heterogeneous annular-sector plate introduced in the preceding sections are selected as representative examples. The distributions of the bending moment components $M_{xx}$, $M_{yy}$, and $M_{xy}$, as well as the transverse shear force components $Q_x$ and $Q_y$, are presented for the median cases of the corresponding test sets and compared with the FEM reference solutions.

Fig. 15 (a) shows the distributions of bending moments and transverse shear forces for the homogeneous single-hole plate under uniform loading. GA-VINO accurately captures the main spatial distribution patterns of all three bending moment components and the transverse shear forces, with discrepancies primarily concentrated around the hole boundary. The relative errors in the shear forces are higher than those in the bending moments. Fig. 15 (b) further presents the results for the annular-sector plate. Similar to the single-hole plate, GA-VINO provides accurate predictions of the three bending moment components. However, substantially larger discrepancies are observed in the transverse shear forces, particularly in the $Q_y$ component, where the predictions deviate noticeably from the FEM reference solution.

This difference may be associated with the relatively small transverse shear deformation in the annular sector plate. According to Mindlin–Reissner plate theory, the transverse shear forces are determined by the difference between the deflection gradients and rotations. When transverse shear deformation is relatively small compared with bending deformation, the deflection gradients and rotations become closer to each other, making the resulting shear strains and shear forces more sensitive to small discrepancies between these predicted quantities. Consequently, accurate predictions of the primary kinematic variables and bending moments do not necessarily guarantee comparable accuracy in the transverse shear forces. These results indicate that GA-VINO provides relatively consistent accuracy in predicting bending moments, whereas accurate recovery of transverse shear forces remains more challenging, particularly when transverse shear deformation is small.

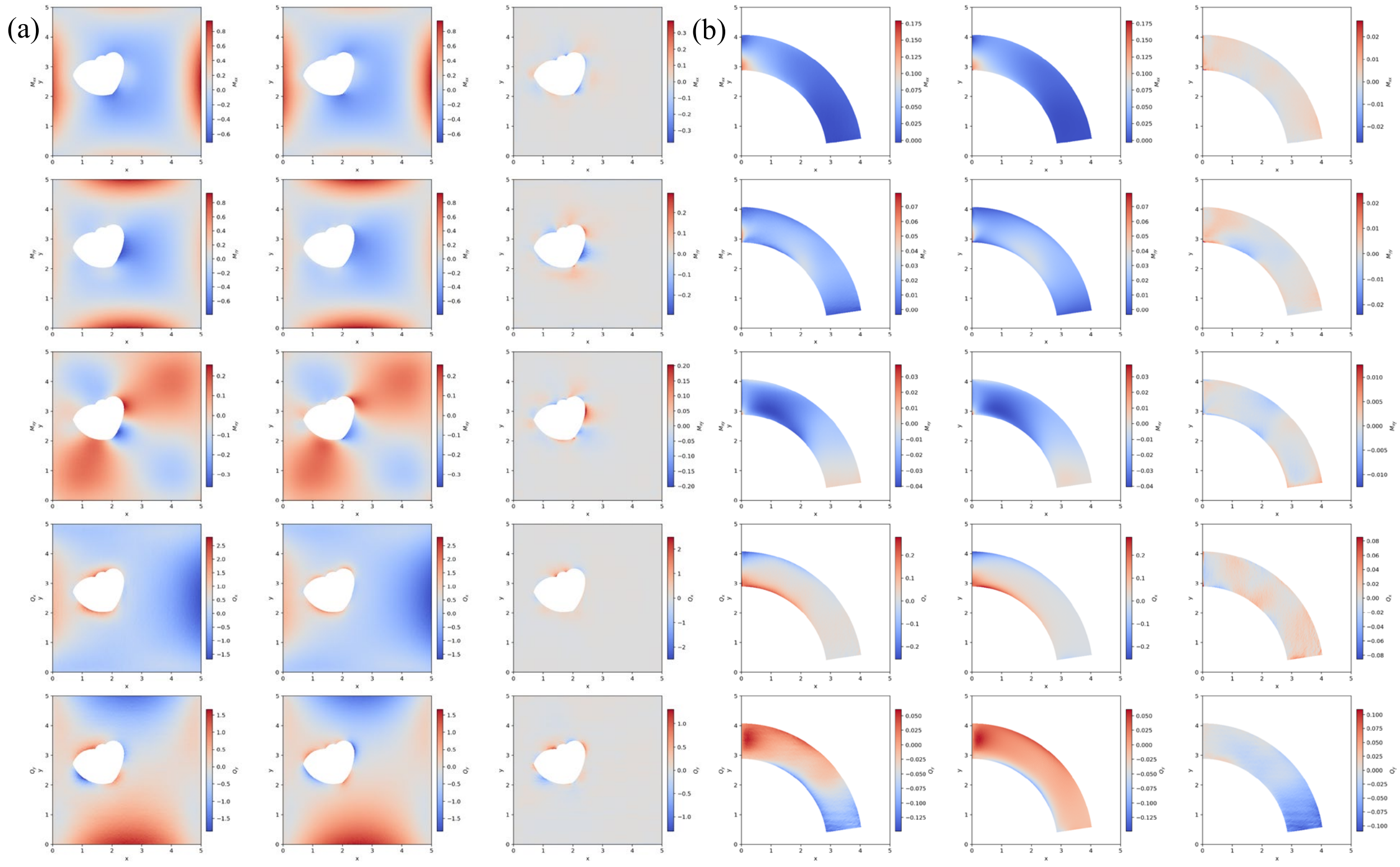

Fig. 15. GA-VINO predictions of bending moments and transverse shear forces and their corresponding error distributions. (a) Median case of the homogeneous single-hole plate ($\|e_{\mathrm{M}}\|_{L^2} = 6.74\%$, $\|e_{\mathrm{Q}}\|_{L^2} = 11.39\%$); (b) median case of the heterogeneous annular-sector plate ($\|e_{\mathrm{M}}\|_{L^2} = 8.80\%$, $\|e_{\mathrm{Q}}\|_{L^2} = 65.31\%$). From top to bottom, the rows show the bending moment components $\mathrm{M}_{xx}$, $\mathrm{M}_{yy}$, and $\mathrm{M}_{xy}$, followed by the transverse shear force components $\mathrm{Q}_x$ and $\mathrm{Q}_y$. From left to right, the columns show the GA-VINO predictions, FEM reference solutions, and corresponding error distributions.

### 4.6. *Out-of-distribution generalization*

In the results presented above, the training and test datasets are sampled from the same data distribution, corresponding to an in-distribution (ID) setting in which the model generalizes to unseen samples within the prescribed geometric and physical parameter ranges. In engineering practice, however, the training distribution may not fully cover all possible scenarios. Therefore, we further investigate the performance of GA-VINO under geometric out-of-distribution (OOD) settings. The hole-plate transfer cases in Sections 4.6.1 and 4.6.2 represent moderate geometric distribution shifts, whereas the L-shaped plate case in Section 4.6.3 represents a stronger geometric extrapolation.

#### 4.6.1. *From double-hole plates to single-hole plates*

As shown in Fig. 16, in this case, GA-VINO is trained using 500 homogeneous double-hole plate samples under uniform loading and is then directly evaluated on 100 homogeneous single-hole plate test samples under uniform loading. For the best case, the relative $L^2$ errors are 0.51% and 1.91%. For the median case, the relative $L^2$ errors are 1.33% and 4.46%, while the worst case has relative $L^2$ errors of 6.43% and 9.97%. Over all test samples, the mean relative $L^2$ errors are 1.73% and 4.08%, with standard deviations of 1.27% and 2.17%. The average $H^1$ seminorm error and energy norm error are 4.27% and 10.23%, respectively.

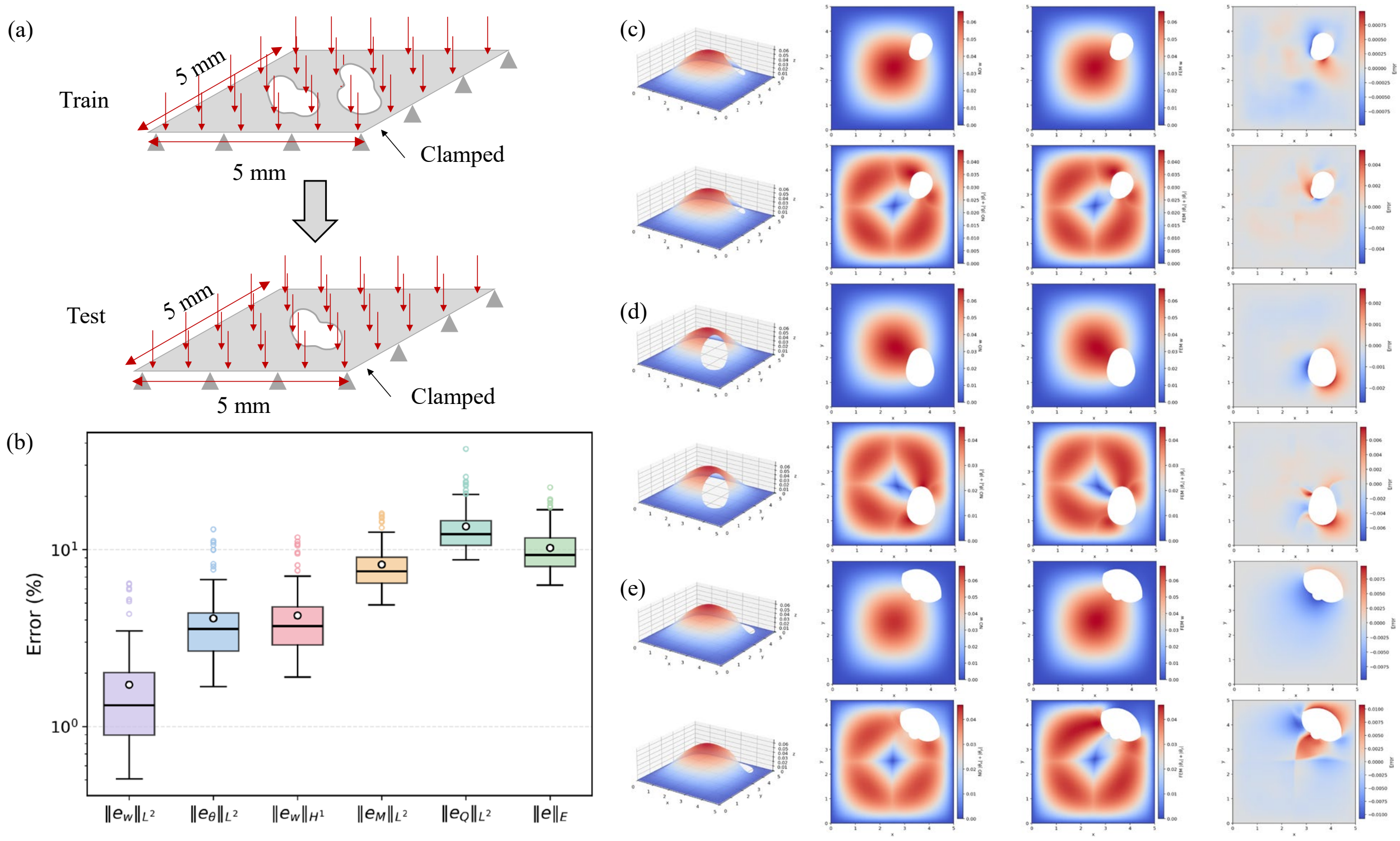

Fig. 16. OOD prediction from double-hole plates to single-hole plates. (a) OOD prediction setting. (b) Relative errors of the predictions. Distributions of the transverse deflection $w$ and the rotation magnitude $|\theta|$ : (c) best case ( $\|e_w\|_{L^2} = 0.51\%$ , $\|e_\theta\|_{L^2} = 1.91\%$ ); (d) median case ( $\|e_w\|_{L^2} = 1.33\%$ , $\|e_\theta\|_{L^2} = 4.46\%$ ); and (e) worst case ( $\|e_w\|_{L^2} = 6.43\%$ , $\|e_\theta\|_{L^2} = 9.97\%$ ). From left to right, the subplots show the surface plot of the transverse deflection, the GA-VINO prediction, the FEM reference solution, and the error.

### *4.6.2. From single-hole plates to double-hole plates*

In this section, GA-VINO is trained using 500 homogeneous single-hole plate samples under uniform loading and is then directly evaluated on 100 double-hole homogeneous plate test samples under uniform loading. For the best case, the relative $L^2$ errors are 0.72% and 2.40%. For the median case, the relative $L^2$ errors are 4.67% and 12.89%, while the worst case has relative $L^2$ errors of 10.67% and 16.03%. Over all test samples, the mean relative $L^2$ errors are 4.93% and 11.55%, with standard deviations of 2.70% and 6.43%. The average $H^1$ seminorm error and energy norm error are 14.22% and 39.17%, respectively. The asymmetric transfer performance may be attributed to the richer geometric diversity of the double-hole training set, which includes overlapping-hole configurations resembling single irregular holes. In contrast, transferring from single-hole to double-hole plates requires the model to capture interactions between holes that are absent from the training data.

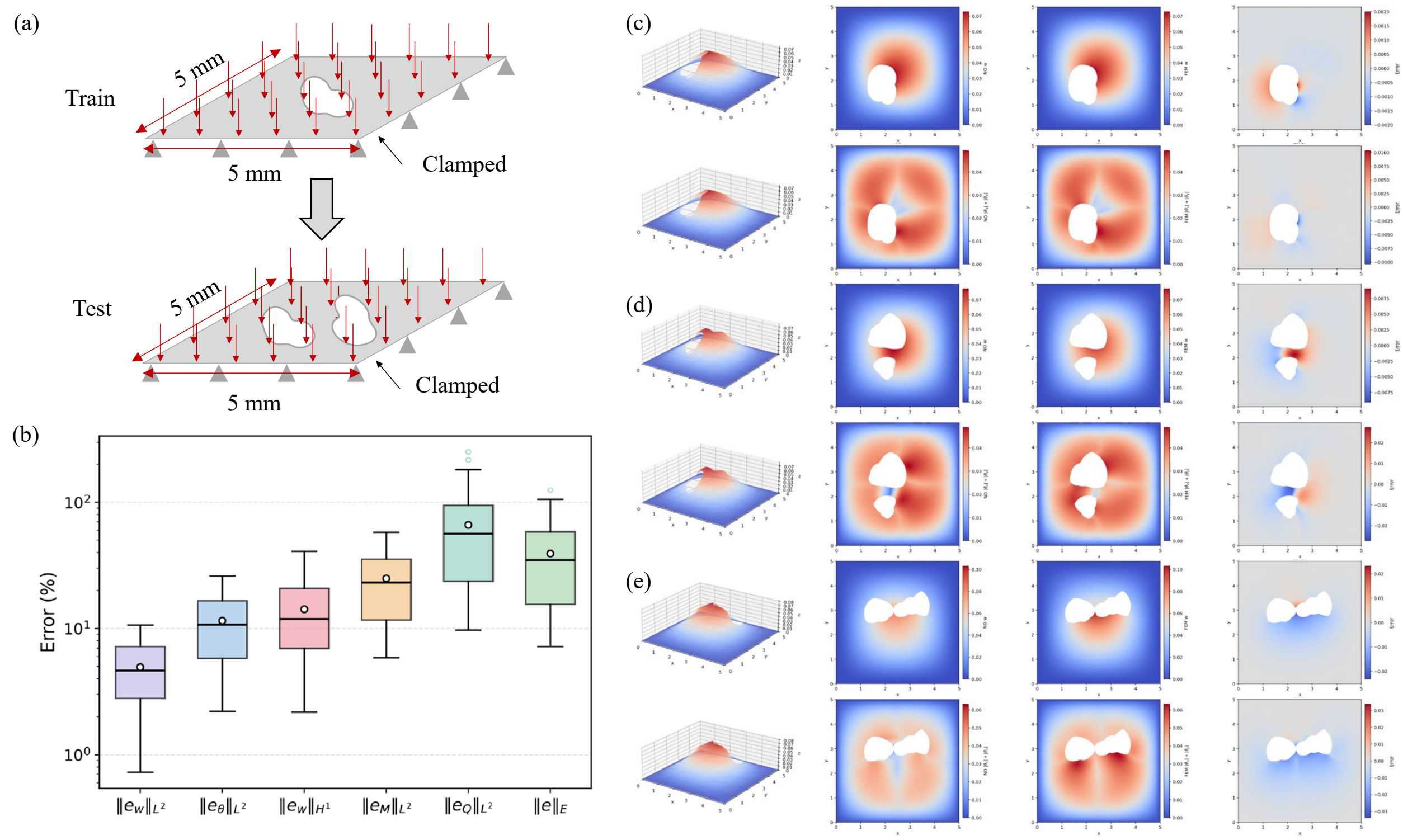

Fig. 17. OOD prediction from single-hole plates to double-hole plates. (a) OOD prediction setting. (b) Relative errors of the predictions. Distributions of the transverse deflection $w$ and the rotation magnitude $|\theta|$ : (c) best case ( $\|e_w\|_{L^2} = 0.72\%$ , $\|e_\theta\|_{L^2} = 2.40\%$ ); (d) median case ( $\|e_w\|_{L^2} = 4.67\%$ , $\|e_\theta\|_{L^2} = 12.89\%$ ); and (e) worst case ( $\|e_w\|_{L^2} = 10.67\%$ , $\|e_\theta\|_{L^2} = 16.03\%$ ). From left to right, the subplots show the surface plot of the transverse deflection, the GA-VINO prediction, the FEM reference solution, and the error.

### *4.6.3. OOD prediction for L-shaped plates*

In this section, we further consider a more challenging geometric OOD problem. In the previous L-shaped plate example, the training geometries were generated by removing a rectangular region from the upper right corner of a square plate, where both the length and width of the rectangular cutout varied from 1.25 mm to 3.25 mm. To evaluate the generalization capability of the model under a stronger geometric distribution shift, the GA-VINO model trained on the L-shaped plates with small cutouts is directly applied to predict L-shaped plates with larger cutouts, where both the length and width of the removed rectangle range from 3.25 mm to 4.25 mm. Meanwhile, the test samples still use spatially uniform loads with sample-dependent amplitudes, and the material parameters are still generated from Gaussian random fields.

The OOD prediction results for this case are shown in Fig. 18. For the best case, the relative $L^2$ errors are 4.65% and 21.64%. For the median case, the relative $L^2$ errors are 15.58% and 19.37%, while the worst case has relative $L^2$ errors of 57.25% and 85.80%. Over all test samples, the mean relative $L^2$ errors are 17.65% and 31.02%, with standard deviations of 10.58% and 12.16%. The average $H^1$ seminorm error and energy norm error are 31.40% and 90.08%, respectively. The larger cutouts substantially alter the effective load-bearing region and deformation patterns, making this extrapolation more challenging than the hole-plate transfer cases. These results highlight the reduced reliability of GA-VINO under pronounced geometric distribution shifts.

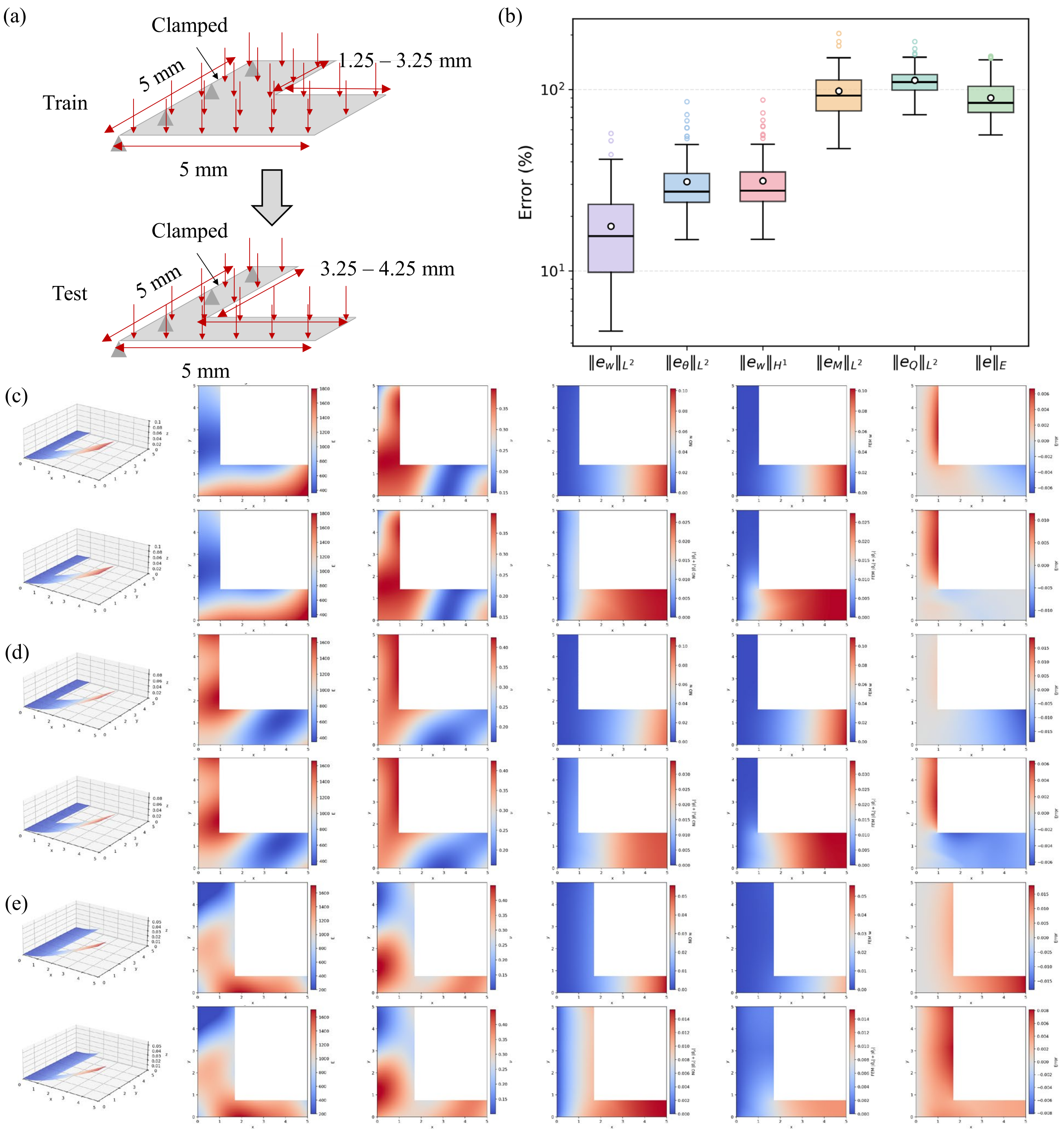

Fig. 18. OOD prediction from small cutout to large cutout L-shaped plates. (a) OOD prediction setting. (b) Relative errors of the predictions. Distributions of the transverse deflection $w$ and the rotation magnitude $|\theta|$: (c) best case ($\|e_w\|_{L^2} = 4.65\%$, $\|e_\theta\|_{L^2} = 21.64\%$); (d) median case ($\|e_w\|_{L^2} = 15.58\%$, $\|e_\theta\|_{L^2} = 19.37\%$); and (e) worst case ($\|e_w\|_{L^2} = 57.25\%$, $\|e_\theta\|_{L^2} = 85.80\%$). From left to right, the subplots show the surface plot of the transverse deflection, the Young's modulus distribution, the Poisson's ratio distribution, the GA-VINO prediction, the FEM reference solution, and the error.

### *4.7. Performance comparison and computational efficiency*

To further evaluate the effectiveness of variational physics-informed training, GA-VINO is compared with both data-driven and strong-form physics-informed neural operators. The data-driven comparisons assess the contribution of incorporating physical constraints, while the strong-form comparisons directly investigate the influence of the physical loss formulation on prediction accuracy and computational efficiency.

#### *4.7.1. Comparison with data-driven neural operators*

To assess the contribution of variational physics-informed training, GA-VINO is compared with purely data-driven neural operators. For the single-hole, double-hole, and annular-sector plate examples, a data-driven counterpart, denoted as GA-NO, is constructed using the same network architecture and input representations as GA-VINO, with the variational energy loss replaced by a supervised mean squared error loss based on FEM solutions. For the L-shaped plate examples, a masked FNO is adopted as a regular-grid-based baseline, where the irregular computational domain is represented within an embedded rectangular grid. The supervised loss used by GA-NO and masked FNO is defined as

$$\mathcal{L} = \frac{\sum_{i=1}^{N_{\text{train}}} \sum_{q=1}^{N_q^{(i)}} m_{i,q} \left\| \mathrm{S}_u^{-1} (\mathbf{u}_{i,q}^{\text{pred}} - \mathbf{u}_{i,q}^{\text{true}}) \right\|_2^2}{3 \sum_{i=1}^{N_{\text{train}}} \sum_{q=1}^{N_q^{(i)}} m_{i,q}}, \tag{40}$$

where $\mathbf{u}_{i,q} = [w_{i,q}, \theta_{x,i,q}, \theta_{y,i,q}]^T$ and $\mathrm{S}_u = \mathrm{diag}(\sigma_w, \sigma_{\theta_x}, \sigma_{\theta_y})$ . Here $\mathbf{u}_{i,q}^{\text{pred}}$ and $\mathbf{u}_{i,q}^{\text{true}}$ denote the neural operator prediction and the FEM reference solution at the $q$-th query point of the $i$-th training sample, respectively. The standard deviations in $\mathrm{S}_u$ are computed from the valid FEM solutions in the training set and are used to balance the contributions of the deflection and rotation components. For GA-NO, $m_{i,q} = 1$ denotes a valid non-padding query point, whereas $m_{i,q} = 0$ denotes a padding point. For masked FNO, $m_{i,q} = 1$ denotes a valid grid point inside the physical domain, whereas $m_{i,q} = 0$ denotes an invalid point in the embedded rectangular grid.

For a fair comparison, the data-driven baselines use the same training and test datasets as the corresponding GA-VINO models, with the main optimization settings kept as consistent as possible. In particular, GA-NO uses the same architecture and training settings as GA-VINO, except for the loss function. For the single-hole and double-hole plate examples, the same early-stopping criterion is applied to both methods. The comparison results are summarized in Table 3, including prediction errors, GPU memory consumption, actual training epochs, and training time.

Table 3 Comparison of GA-VINO with data-driven neural operators. The reported error metrics are averaged over all samples in the corresponding test set.

| Example/setting | Baseline | $\|e_w\|_{L^2}$ (%) | $\|e_\theta\|_{L^2}$ (%) | $\|e_w\|_{H^1}$ (%) | $\|e_M\|_{L^2}$ (%) | $\|e_Q\|_{L^2}$ (%) | $\|e\|_E$ (%) | Memory (MB) | Actual epochs | Training time (s) |
|---|---|---|---|---|---|---|---|---|---|---|
| Single-hole plate (UM, UL) | GA-VINO | 2.13 | 4.11 | 4.57 | 8.04 | 16.53 | 11.09 | 12917.35 | 234 | 1,412.90 |
| | GA-NO | 1.53 | 3.76 | 8.39 | 12.10 | 44.13 | 23.59 | 12,930.66 | 284 | 2,192.51 |
| Single-hole plate (UM, RL) | GA-VINO | 2.27 | 3.88 | 4.06 | 6.50 | 11.14 | 8.25 | 11,049.73 | 326 | 4,332.44 |
| | GA-NO | 2.63 | 4.94 | 7.32 | 11.09 | 33.37 | 18.99 | 11,261.57 | 399 | 6,079.18 |
| Double-hole plate (UM, UL) | GA-VINO | 2.94 | 6.28 | 6.51 | 12.44 | 20.77 | 15.39 | 12,127.06 | 219 | 1,170.87 |
| | GA-NO | 2.91 | 6.82 | 16.60 | 23.46 | 90.65 | 47.78 | 12,140.18 | 165 | 1,179.12 |
| Double-hole plate (UM, RL) | GA-VINO | 3.98 | 6.97 | 7.10 | 11.49 | 17.37 | 13.69 | 10,600.46 | 393 | 4,323.47 |
| | GA-NO | 4.38 | 8.49 | 13.04 | 20.39 | 60.58 | 34.44 | 10,608.98 | 216 | 3,003.39 |
| L-shaped plate (RM, UL) | GA-VINO | 2.23 | 3.37 | 3.47 | 8.90 | 44.49 | 12.17 | 10,439.34 | 500 | 3,833.33 |
| | masked FNO | 1.44 | 1.60 | 22.58 | 24.64 | 1485.61 | 244.58 | 919.47 | 500 | 316.42 |
| L-shaped plate (RM, RL) | GA-VINO | 3.61 | 4.96 | 5.00 | 9.78 | 46.14 | 13.28 | 12,618.92 | 500 | 5,038.92 |
| | masked FNO | 1.92 | 2.32 | 27.13 | 30.72 | 1,681.76 | 288.18 | 919.47 | 500 | 324.23 |
| Annular sector (RM, RL) | GA-VINO | 3.93 | 4.62 | 5.15 | 9.29 | 59.51 | 16.21 | 7,397.54 | 500 | 5,438.75 |
| | GA-NO | 6.80 | 6.40 | 15.01 | 18.60 | 589.81 | 119.41 | 7,405.21 | 500 | 6,873.42 |
| OOD from double-hole to single-hole plates | GA-VINO | 1.73 | 4.08 | 4.27 | 8.24 | 13.51 | 10.23 | / | / | / |
| | GA-NO | 2.54 | 5.73 | 17.09 | 20.86 | 92.29 | 47.33 | / | / | / |
| OOD from single-hole plates to double-hole plates | GA-VINO | 4.93 | 11.55 | 14.22 | 24.99 | 66.39 | 39.17 | / | / | / |
| | GA-NO | 5.23 | 12.16 | 17.51 | 30.84 | 87.28 | 49.81 | / | / | / |
| OOD prediction for L-shaped plates | GA-VINO | 17.65 | 31.02 | 31.40 | 98.07 | 112.40 | 90.08 | / | / | / |
| | masked FNO | 23.19 | 27.87 | 74.77 | 102.01 | 3,535.41 | 708.18 | / | / | / |

As shown in Table 3, the data-driven baselines achieve comparable or even lower $L_2$ errors for the primary variables in some cases, but generally exhibit larger errors in gradient-sensitive and energy-based metrics, particularly for bending moments and transverse shear forces. This difference can be attributed to the variational energy constraint in GA-VINO, which explicitly incorporates bending and transverse shear contributions, whereas supervised MSE training mainly minimizes pointwise prediction errors. For the L-shaped plates, masked FNO achieves lower errors in deflection and rotations with substantially shorter training times and lower GPU memory consumption, but exhibits much larger errors in the derived mechanical quantities. These results indicate that pointwise accuracy alone is insufficient to assess the mechanical consistency of predicted response fields.

### *4.7.2. Comparison with strong-form PINOs*

To further investigate the advantages of the variational formulation over strong-form physics-informed training, the homogeneous single-hole plate under uniform loading introduced in Section 4.1.1 is selected as a representative benchmark. A strong-form PINO is constructed based on the governing equations of Mindlin–Reissner plates and compared with GA-VINO. To minimize the influence of architectural differences, the strong-form PINO and GA-VINO employ the same geometry-aware neural operator architecture and primary output variables, with the main distinction being the formulation of the physics-informed loss function.

To separately investigate the effects of purely physics-driven training and data-assisted training, the total loss function is defined as

$$L = L_{\text{pde}} + \lambda_{\text{data}} L_{\text{data}} \tag{41}$$

where $L_{\text{pde}}$ denotes the physics-informed loss, formulated as either the variational energy loss $L_{\text{pde-v}}$ or the strong-form loss $L_{\text{pde-s}}$, depending on the training formulation. Here, $L_{\text{data}}$ denotes the supervised data loss, and $\lambda_{\text{data}}$ is its corresponding weight. For purely physics-driven training, $\lambda_{\text{data}} = 0$. For data-assisted training, $\lambda_{\text{data}}$ is set to $10^{-4}$ and $10^{3}$ for GA-VINO + data and PINO + data, respectively.

The strong-form physics-informed loss comprises four contributions associated with transverse force equilibrium, moment equilibrium, free-boundary shear force, and free-boundary bending moment, and is defined as

$$L_{\text{pde-s}} = \frac{1}{N_{\text{train}}} \sum_{i=1}^{N_{\text{train}}} [\lambda_{\text{force}} L_{\text{force}}^{(i)} + \lambda_{\text{rotation}} L_{\text{rotation}}^{(i)} + \lambda_{\text{free-shear}} L_{\text{free-shear}}^{(i)} + \lambda_{\text{free-moment}} L_{\text{free-moment}}^{(i)}] \tag{42}$$

For the interior of the computational domain, the residual losses associated with the equilibrium equations are defined as

$$L_{\text{force}}^{(i)} = \frac{1}{N_r} \sum_{j=1}^{N_r} [\nabla \cdot \mathrm{Q}^{(i)}(\mathbf{x}_{r,j}^{(i)}) + p^{(i)}(\mathbf{x}_{r,j}^{(i)})]^2 \tag{43}$$

$$L_{\text{rotation}}^{(i)} = \frac{1}{N_r} \sum_{j=1}^{N_r} \left\| \nabla \cdot \mathrm{M}^{(i)}(\mathbf{x}_{r,j}^{(i)}) + \mathrm{Q}^{(i)}(\mathbf{x}_{r,j}^{(i)}) \right\|_2^2 \tag{44}$$

The residual losses associated with the free boundary conditions on the free boundaries are expressed as

$$L_{\text{free-shear}}^{(i)} = \frac{1}{N_b} \sum_{j=1}^{N_b} [\mathrm{Q}^{(i)}(\mathbf{x}_{b,j}^{(i)}) \cdot n_j^{(i)}]^2, \tag{45}$$

$$L_{\text{free-moment}}^{(i)} = \frac{1}{N_b} \sum_{j=1}^{N_b} \left\| \mathrm{M}^{(i)}(\mathbf{x}_{b,j}^{(i)}) n_j^{(i)} \right\|_2^2. \tag{46}$$

Here, $\mathbf{x}_{r,j}^{(i)}$ and $\mathbf{x}_{b,j}^{(i)}$ denote the $j$-th interior and free-boundary collocation points of the $i$-th training sample, respectively, and $\mathbf{n}_j^{(i)}$ is the outward unit normal vector at the corresponding boundary point. $N_r$ and $N_b$ denote the numbers of interior and free-boundary collocation points per sample, respectively. All loss components are non-dimensionalized. Considering the differences in numerical magnitudes and optimization sensitivities among the physical residuals, separate weights are assigned to the individual loss components. In this benchmark, the weights are set to $\lambda_{\text{force}} = 1$, $\lambda_{\text{rotation}} = 100$, $\lambda_{\text{free-shear}} = 100$, $\lambda_{\text{free-moment}} = 10^7$.

For the clamped conditions on the outer boundary, the strong-form model employs a continuous output transformation to impose hard constraints, thereby avoiding an additional boundary penalty term. Given the raw network output $\mathbf{u}_{\Theta}^{\text{raw}}$, the final prediction is defined as

$$\mathbf{u}_{\Theta}(\mathbf{x}_q) = \phi(\xi, \zeta) \mathbf{u}_{\Theta}^{\text{raw}}(\mathbf{x}_q) \tag{47}$$

where $\phi(\xi,\zeta) = (1-\xi^2)(1-\zeta^2)$, and $\xi, \zeta \in [-1,1]$ are the normalized coordinates obtained by linearly mapping the physical coordinates of the plate. This transformation ensures that the transverse deflection and both rotations vanish exactly along the outer boundary, thereby satisfying the clamped boundary conditions.

To improve the stability of strong-form training, the initial learning rate is set to $10^{-4}$, while the network architecture and other training hyperparameters are kept as consistent as possible with those of GA-VINO. All models use the same single-hole plate dataset and are evaluated on the same test set. Each model is trained for a maximum of 500 epochs. For the strong-form PINO, the numbers of interior and free-boundary collocation points are set to $N_r = 512$ and $N_b = 128$, respectively.

Table 4 Comparison of prediction accuracy and computational costs among different training formulations.

| Baseline | $\|e_w\|_{L^2}$ (%) | $\|e_\theta\|_{L^2}$ (%) | $\|e_w\|_{H^1}$ (%) | $\|e_{\mathrm{M}}\|_{L^2}$ (%) | $\|e_{\mathrm{Q}}\|_{L^2}$ (%) | $\|e\|_E$ (%) | Training time (s) |
|---|---|---|---|---|---|---|---|
| GA-VINO | 1.43 | 3.31 | 3.75 | 6.99 | 14.16 | 9.63 | 3,019.00 |
| GA-NO | 1.53 | 3.76 | 8.39 | 12.10 | 44.13 | 23.59 | 3,860.05 |
| GA-VINO + data | 1.38 | 3.04 | 3.33 | 6.22 | 11.59 | 8.17 | 3,023.35 |
| Strong-form | 89.46 | 95.76 | 92.42 | 92.83 | 66.63 | 88.00 | 5,782.05 |
| Strong-form + data | 7.15 | 25.66 | 22.51 | 41.12 | 66.14 | 49.66 | 5,816.95 |

As shown in Table 4, the purely physics-driven strong-form PINO exhibits substantially larger prediction errors than GA-VINO. Incorporating supervised data improves the accuracy of both formulations, although data-assisted training requires labeled FEM solutions and introduces additional offline data-generation costs. Even with supervised data, the strong-form PINO remains less accurate than GA-VINO and GA-VINO + data across the evaluated metrics. The strong-form residuals involve second-order spatial derivatives and separately weighted equilibrium and boundary terms, which may increase computational overhead and optimization difficulty. In contrast, GA-VINO employs lower-order derivatives and element-wise energy evaluation, providing a more efficient training objective. Overall, these results demonstrate the advantages of variational physics-informed training in terms of prediction accuracy and training efficiency for the Mindlin–Reissner plate benchmark considered here, while the performance of the strong-form approach may depend on residual weighting and optimization settings.

#### *4.7.3. Computational cost and amortization analysis*

Although GA-VINO achieves substantially faster inference than FEM, its offline training cost should also be considered when evaluating the overall computational efficiency. For applications requiring repeated simulations under varying geometries, material properties, and loading conditions, the computational advantage depends on the number of new problems to be solved after training. To quantify this trade-off, we consider the total computational costs of GA-VINO and FEM for $N$ new simulations

$$T_{\mathrm{NO}}^{\mathrm{total}}(N) = T_{\mathrm{train}} + NT_{\mathrm{infer}} \tag{48}$$

$$T_{\mathrm{FEM}}^{\mathrm{total}}(N) = NT_{\mathrm{FEM}} \tag{49}$$

where $T_{\mathrm{train}}$ denotes the offline training time of GA-VINO, and $T_{\mathrm{infer}}$ and $T_{\mathrm{FEM}}$ represent the time required for a single GA-VINO inference and FEM solution, respectively. Accordingly, the break-even number of simulations, beyond which the total computational cost of GA-VINO becomes lower than that of FEM, is given by

$$N_{\text{break-even}} = \frac{T_{\mathrm{train}}}{T_{\mathrm{FEM}} - T_{\mathrm{infer}}} \tag{50}$$

Using the computational times reported in Table 2 and Table 3, the estimated break-even numbers are approximately 201, 696, and 2,248 for the single-hole plate under uniform loading, the double-hole plate under random loading, and the annular-sector plate, respectively. These results indicate that the substantial online acceleration of GA-VINO can offset its initial training cost when a sufficient number of new simulations are required. However, the number of simulations needed for amortization varies across the considered problems, depending on the training cost and the computational expense of the corresponding FEM solutions.

### *4.8. Robustness to training randomness*

To assess the robustness of the proposed method to randomness in neural-network training, repeated training experiments were conducted on two representative benchmark problems: the single-hole plate (UM, UL) and double-hole plate (UM, RL) cases. For each case, GA-VINO and GA-NO were each trained independently five times using different random seeds.

In each independent run, the random seed controls the network parameter initialization, the randomized ordering of training samples, and the initial point used in FPS for geometric point-cloud encoding. To ensure a fair comparison, GA-VINO and GA-NO with the same random seed use the same data split, training-sample ordering, and FPS initialization condition. Apart from these seed-controlled factors, all other settings are identical to those used in the corresponding benchmark cases described above. For each run, relative errors were first evaluated over the fixed test set, and the mean error over the test samples was obtained. The mean and standard deviation were then calculated from the corresponding results of the five independent runs. Table 5 summarizes the errors for the different cases. Notably, the mean and standard deviation reported in the table represent the statistics across the five independent random seeds, rather than the variation among test samples within a single run.

Overall, both methods exhibit small performance variations across different random seeds in the two benchmark cases. These results indicate that GA-VINO is robust to training randomness, and that the observed accuracy improvements are not attributable to a favorable initialization or optimization trajectory in a single run.

Table 5 Means and standard deviations of relative errors across five independent training runs with different random seeds.

| Example/setting | Baseline | $\|e_w\|_{L^2}$ (%) (mean/std) | $\|e_\theta\|_{L^2}$ (%) (mean/std) | $\|e_w\|_{H^1}$ (%) (mean/std) | $\|e_M\|_{L^2}$ (%) (mean/std) | $\|e_Q\|_{L^2}$ (%) (mean/std) | $\|e\|_E$ (%) (mean/std) |
|---|---|---|---|---|---|---|---|
| Single-hole plate (UM, UL) | GA-VINO | 1.74/0.16 | 3.69/0.23 | 4.06/0.27 | 7.29/0.47 | 14.89/0.78 | 10.03/0.59 |
| | GA-NO | 1.48/0.08 | 3.58/0.23 | 7.71/1.06 | 11.47/1.26 | 40.02/6.61 | 21.64/3.23 |
| Double-hole plate (UM, RL) | GA-VINO | 4.13/0.40 | 6.84/0.36 | 6.91/0.38 | 11.09/0.57 | 17.20/1.15 | 13.35/0.76 |
| | GA-NO | 4.28/0.16 | 8.25/0.28 | 12.75/0.14 | 19.25/0.72 | 59.97/1.77 | 33.63/0.65 |

### 4.9. Sensitivity to discretization

As the variational loss is evaluated through element-wise integration over a triangular discretization of the computational domain, the discretization used during training may affect the numerical evaluation of the energy functional and, consequently, the learned solution. To assess the robustness of GA-VINO with respect to the training discretization, two complementary aspects are investigated in this section: discretization density and element quality.

Both studies are conducted using the homogeneous single-hole plate under uniform loading introduced in Section 4.1.1. For each discretization considered in this section, GA-VINO is trained for the full 500 epochs, while the network architecture, optimization settings, and all other training hyperparameters are kept unchanged. In the discretization-density study, the boundary and query point clouds are progressively coarsened. In the element-quality study, meshes with progressively degraded triangular-element quality are considered while the total number of points is kept approximately unchanged.

#### 4.9.1. Effect of discretization density

To investigate the influence of discretization density on the prediction performance of GA-VINO, five discretization levels with progressively reduced geometry point-cloud sizes $N_g$ and numbers of query points $N_q$ are considered, as summarized in Table 6.

To isolate the influence of the training discretization from the numerical error of the FEM reference solution, the reference solutions for all discretization levels are obtained using a sufficiently refined FEM mesh and are then interpolated onto the corresponding query points. Accordingly, the relative $L_2$ errors reported in this study are evaluated at the query points associated with each discretization level, enabling a consistent comparison across different point-cloud resolutions.

Table 6 Influence of input point cloud and query point densities on the prediction performance and training efficiency of GA-VINO.

| $N_g$ | $N_q$ | $\|e_w\|_{L^2}$ (%) | $\|e_\theta\|_{L^2}$ (%) | Memory (MB) | Training time (s) |
|---|---|---|---|---|---|
| [168-180] | [1,224-2,000] | 13.37 | 12.31 | 1,173.57 | 708.77 |
| [204-232] | [2,630-4,732] | 3.73 | 5.15 | 2,094.01 | 899.55 |
| [260-292] | [2,688-4,887] | 2.23 | 4.06 | 3,700.29 | 1,184.56 |
| [360-396] | [5,376-9,698] | 2.30 | 5.43 | 6,878.13 | 1,801.22 |
| 528 | [10,523-18,821] | 1.43 | 3.31 | 12,917.35 | 3,019.01 |

Overall, the results show a clear trend of decreasing prediction accuracy as the discretization density is reduced, which is consistent with the observations reported for VINO by Eshaghi et al. [32]. However, a strictly monotonic variation in error is not observed between every two successive discretization levels. This is reasonable because $N_g$ and $N_q$ are not the only discretization-related factors affecting the model performance. Other sampling and grouping parameters, such as $N_s$ and $N_p$, are kept fixed throughout this study and may influence the relative performance at intermediate resolutions. Nevertheless, once the point clouds become sufficiently sparse, the prediction errors increase markedly.

The results also demonstrate that GA-VINO is reasonably robust to moderate reductions in discretization density.

Therefore, an appropriately coarsened discretization can substantially reduce GPU memory consumption and training cost while incurring only a limited loss of accuracy. Excessively sparse discretizations should nevertheless be avoided, as insufficient spatial resolution can limit the model's ability to represent both geometric features and solution fields.

#### *4.9.2. Effect of triangular-element quality*

In addition to discretization density, the triangular discretization used for evaluating the variational loss may also influence the learned solution. To investigate this effect, several discretizations with different element qualities are generated while keeping the overall number of query points approximately unchanged. Across these discretizations, both the nodal positions and triangular connectivities vary, allowing the sensitivity of GA-VINO to different training discretizations to be assessed at approximately the same resolution.

The quality of an individual triangular element is quantified by

$$q_e = \frac{4\sqrt{3}A_e}{l_1^2 + l_2^2 + l_3^2} \tag{51}$$

where $A_e$ is the area of the $e$-th triangular element, and $l_1$, $l_2$, and $l_3$ denote its three edge lengths. The metric satisfies $0 < q_e \le 1$, with $q_e = 1$ corresponding to an equilateral triangle and smaller values indicating increasingly distorted elements.

To characterize the lower tail of the triangular-element quality distribution, the 5th percentile of $q_e$ over all triangular elements, denoted by $q_{5\%}$, is adopted as the mesh-level quality indicator. A smaller $q_{5\%}$ indicates more severe element distortion in the lower-quality portion of the mesh.

Table 7 summarizes the prediction errors obtained with different values of $q_{5\%}$. Since the evaluation relies on FEM reference solutions, the robustness of the adopted FEM formulation with respect to element distortion is independently examined in Appendix D using a problem with an analytical solution. The results in Appendix D show that the FEM solution remains stable over a broad range of triangular-element qualities, supporting its use as the reference solution in the present comparison.

Table 7 Influence of triangular-element quality on the prediction performance of GA-VINO.

| $q_{5\%}$ | $\|e_w\|_{L^2}$ (%) | $\|e_\theta\|_{L^2}$ (%) | $\|e_w\|_{H^1}$ (%) | $\|e_M\|_{L^2}$ (%) | $\|e_Q\|_{L^2}$ (%) | $\|e\|_E$ (%) |
|---|---|---|---|---|---|---|
| 0.9486 | 1.43 | 3.31 | 3.75 | 6.99 | 14.16 | 9.63 |
| 0.6917 | 1.37 | 3.01 | 3.30 | 6.08 | 11.21 | 8.02 |
| 0.4395 | 1.36 | 3.02 | 3.38 | 6.39 | 12.03 | 8.48 |
| 0.1884 | 1.63 | 3.50 | 3.86 | 7.33 | 13.54 | 9.64 |

As shown in Table 7, the prediction errors remain relatively stable across the considered triangular discretizations. Although slight fluctuations are observed as the element-quality indicator $q_{5\%}$ decreases, no systematic deterioration in prediction accuracy is observed, and the lowest-quality discretization still yields errors comparable to those obtained with higher-quality discretizations. Overall, these results indicate that GA-VINO maintains relatively consistent prediction accuracy under the investigated variations in triangular-element quality, nodal positions, and element connectivity.

### *4.10. Thickness sensitivity*

All preceding examples employ a variational loss based on Mindlin–Reissner plate theory, with the plate thickness kept constant. However, previous studies [16][34] have shown that, for thin plate and shell structures, the bending and transverse shear stiffnesses scale with $h^3$ and $h$, respectively. As the thickness decreases, the increasing disparity between these stiffness scales may lead to ill-conditioned optimization in energy-based physics-informed neural networks, resulting in slower training convergence or even shear locking. For Mindlin–Reissner plates, the transverse shear strain $\gamma = \nabla w - \theta$ should approach zero in the thin-plate limit. If the numerical discretization cannot adequately satisfy this constraint, nonphysical residual shear strains may arise, potentially leading to shear locking.

To investigate the effect of plate thickness on the predictive performance of GA-VINO, we consider the homogeneous single-hole plate benchmark introduced in Section 4.1.1. Five thickness-to-length ratios are examined, $h/L$ = $10^{-1}$, $10^{-1.5}$, $10^{-2}$, $10^{-2.5}$, $10^{-3}$, corresponding to plate thicknesses of $h$ = 0.5 mm, 0.158 mm, 0.05 mm, 0.0158 mm and 0.005 mm, respectively, with $L = 5$ mm. To maintain comparable magnitudes of transverse deflection across different thicknesses, the uniformly distributed pressure is scaled according to $p \propto h^3$, while all other model configurations and training settings remain unchanged.

We first train GA-VINO using the original variational loss based on Mindlin–Reissner plate theory. As shown in Fig. 19, GA-VINO achieves high accuracy in predicting both transverse deflection and rotations at relatively large thickness-to-length ratios. However, the prediction errors increase as the plate thickness decreases, with a pronounced deterioration observed when $h/L=10^{-2}$. For the thinnest plate, the model fails to produce accurate predictions. This trend suggests that the current low-order MR discretization becomes increasingly affected by shear locking and the associated ill-conditioning of energy-based optimization in the thin-plate regime.

To further investigate the applicability of GA-VINO to thin plates, the variational loss is reformulated based on Kirchhoff–Love plate theory. Unlike Mindlin–Reissner theory, Kirchhoff–Love theory neglects transverse shear deformation and satisfies $\gamma=\nabla w-\theta=0$. Consequently, the transverse deflection $w$ is taken as the only primary unknown, while the rotations are obtained from its first-order spatial derivatives. For the KL plate considered in this example, the clamped boundary conditions are expressed as

$$w=0,\ \frac{\partial w}{\partial n}=0, (\text{at x=0, x=5, y=0, y=5}) \tag{52}$$

These clamped boundary conditions are imposed through a smooth hard-output transformation, similar to that used in Eq. (47). Given the raw network output $w_{\Theta}^{\text{raw}}$, the constrained transverse deflection is defined as

$$w_{\Theta}(\mathbf{x}_q)=\phi^2(\xi,\zeta)w_{\Theta}^{\text{raw}}(\mathbf{x}_q). \tag{53}$$

Since both $\phi^2$ and its first-order spatial derivatives vanish along the outer boundary, the clamped boundary conditions are satisfied exactly without additional boundary penalty terms.

The bending curvature tensor is expressed in terms of the second-order spatial derivatives of the transverse deflection as

$$\kappa_{\text{KL}}=\nabla^2 w. \tag{54}$$

The corresponding bending strain energy density is given by

$$\psi_{\text{bend}}^{\text{KL}}=\frac{1}{2}\frac{h^3}{12}\left[\lambda\left(\text{tr}\kappa_{\text{KL}}\right)^2+2\mu\left(\kappa_{\text{KL}}:\kappa_{\text{KL}}\right)\right]. \tag{55}$$

The total potential energy can thus be written as

$$\Pi_{\text{KL}}=\int_{\Omega}[\psi_{\text{bend}}^{\text{KL}}-pw]d\Omega=U_{\text{bend}}^{\text{KL}}-W_{\text{ext}}. \tag{56}$$

For the Kirchhoff–Love formulation, the curvature tensor is obtained directly from the second-order spatial derivatives of the boundary-constrained neural network output using automatic differentiation. For each triangular element, the curvature tensors evaluated at its three vertices are averaged, and the bending energy is approximated by multiplying the corresponding energy density by the element area and summing over all elements.

As shown in Fig. 19, reformulating the variational loss based on Kirchhoff–Love plate theory significantly improves the predictive accuracy of GA-VINO at smaller thickness-to-length ratios. When $h/L$ increases from $10^{-3}$ to $10^{-2}$, the relative error in transverse deflection remains approximately 6%–7%, while that in rotations stays around 12%. In contrast, at $h/L=10^{-1}$, the corresponding errors increase to 24.98% and 16.60%, respectively. This behavior is primarily attributed to the kinematic assumptions of Kirchhoff–Love theory, which neglects transverse shear deformation. In the thin-plate regime, this assumption avoids the locking issue associated with the zero-shear-strain constraint in the original Mindlin–Reissner discretization, thereby improving predictive accuracy. However, as the plate thickness increases, transverse shear deformation becomes increasingly significant, and the modeling approximations inherent in Kirchhoff–Love theory may lead to larger prediction errors. Furthermore, since the rotations in Kirchhoff–Love theory are obtained from the first-order spatial derivatives of the predicted transverse deflection, their errors are generally higher than those of the deflection in the thin-plate regime.

Overall, the Mindlin–Reissner-based variational formulation employed in this study achieves high predictive accuracy for moderately thick plates. However, as the thickness-to-length ratio decreases, shear locking associated with the low-order discretization and the resulting optimization difficulties may lead to a substantial deterioration in accuracy. To extend the applicability of GA-VINO to thin plates, the Kirchhoff–Love-based variational formulation can serve as a complementary approach to the original Mindlin–Reissner formulation. Based on the present results, $h/L=10^{-2}$ may be tentatively considered as a reference threshold for switching between the two formulations, although further numerical investigations are required to validate this criterion. It should be noted that, despite the considerable improvement achieved with Kirchhoff–Love theory in the thin-plate regime, the predictive accuracy remains less than satisfactory. Future work will focus on developing improved variational optimization strategies for thin plates to enhance the accuracy and stability of GA-VINO over a broader range of thickness-to-length ratios.

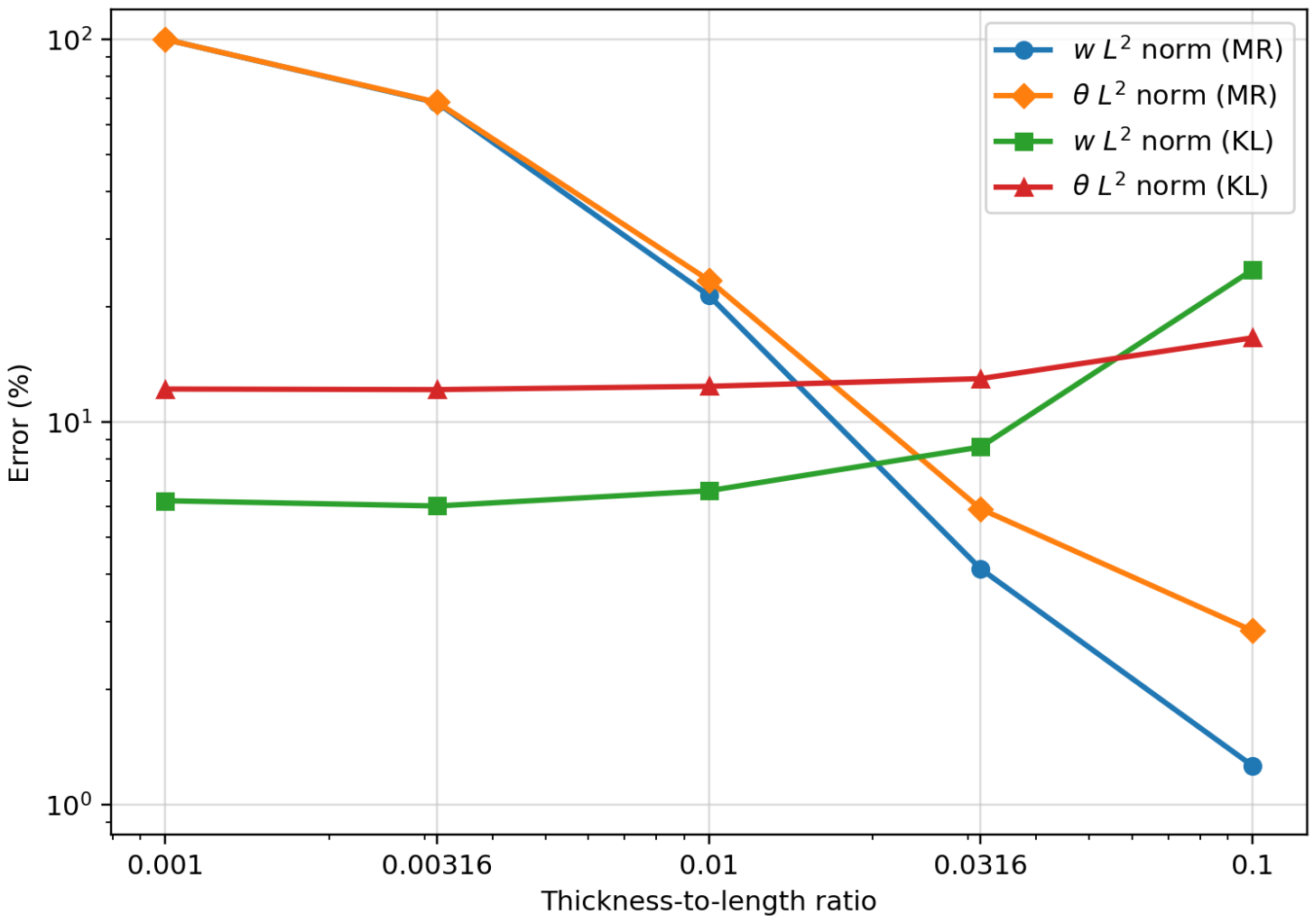

Fig. 19. Mean relative $L^2$ errors in transverse deflection and rotation for the Mindlin–Reissner (MR) and Kirchhoff–Love (KL) formulations at different thickness-to-length ratios.

## 5. Conclusions

In this work, we proposed a geometry-aware variational physics-informed neural operator (GA-VINO) for Mindlin–Reissner plates with complex geometries and varying physical parameters. The framework employs a point-cloud-based encoder–decoder architecture, where boundary point clouds are used to represent the computational geometry. Dedicated material and load encoders, together with a scalar-parameter branch, are flexibly incorporated to handle heterogeneous material fields, spatially varying loads, and sample-level scalar parameters within a unified framework. Through multi-branch feature fusion and cross-attention mechanisms, GA-VINO integrates geometric, material, loading, and query-point information to directly predict transverse deflection and rotations at arbitrary query locations. Compared with neural operators that rely on regular grids or fixed computational domain representations, GA-VINO offers greater flexibility in handling irregular geometries and allows different physical inputs to be discretized on different point sets, avoiding the need to interpolate all physical fields onto a common grid. More importantly, GA-VINO is trained in a purely physics-driven manner by minimizing a variational loss constructed from the discretized total potential energy of Mindlin–Reissner plates, without requiring labeled high-fidelity solutions. This substantially reduces the dependence on supervised data associated with conventional data-driven neural operators.

Numerical results demonstrate that GA-VINO accurately predicts the primary kinematic responses across different plate geometries, heterogeneous materials, and loading conditions. Compared with purely data-driven neural operators, GA-VINO generally achieves lower gradient-related and energy-norm errors, even when the prediction accuracies of the primary kinematic variables are comparable. In particular, it shows clear advantages in predicting derived mechanical quantities such as bending moments and transverse shear forces. For a representative benchmark, GA-VINO also achieves higher prediction accuracy than the strong-form PINO under the investigated training settings. Nevertheless, the prediction errors of transverse shear forces remain considerably larger than those of bending moments in some cases, indicating that accurately recovering local stress resultants from the predicted kinematic fields remains challenging.

The robustness studies show that GA-VINO exhibits limited performance variations across five independent training runs with different random seeds and maintains relatively stable prediction accuracy under the investigated variations in triangular-element quality. Although reducing the training discretization density can decrease computational costs, excessively coarse discretizations lead to a noticeable deterioration in prediction accuracy. In terms of computational efficiency, GA-VINO achieves millisecond-level full-field inference, providing online speedups of approximately two to three orders of magnitude compared with the GPU-accelerated FEM solver. Its offline training cost can be effectively amortized when a sufficiently large number of new simulations are required. Furthermore, GA-VINO demonstrates useful generalization capability under moderate geometric distribution shifts, although its prediction accuracy decreases as the degree of geometric extrapolation increases.

Despite these advantages, several limitations remain. First, the current framework is based on linear elasticity and small-deformation Mindlin–Reissner plate theory, without explicitly accounting for membrane strain energy or geometric nonlinearities. Future extensions may incorporate membrane energy, nonlinear strain–displacement relations, and configuration-update strategies to address geometrically nonlinear plate and shell problems. In addition, the low-order Mindlin–Reissner discretization encounters increasing numerical difficulties in the thin-plate regime, where shear locking and associated optimization difficulties may substantially affect prediction accuracy. Reformu-lating the variational loss based on Kirchhoff–Love plate theory improves the predictions for thin plates, but its applicability becomes limited as transverse shear deformation becomes more significant. Therefore, future work will investigate locking-free variational discretization and optimization strategies to improve the accuracy and stability of GA-VINO over a broader range of thickness-to-length ratios. To further improve the prediction of derivative-related mechanical quantities, Sobolev training strategies [35][36] may be incorporated. Moreover, to address the deterioration in prediction accuracy under larger geometric distribution shifts, future research may explore fine-tuning with a limited number of target-domain samples [37], as well as neural operator pretraining and warm-start strategies [38][39][40], to enhance the adaptability of GA-VINO to new geometric configurations and physical parameter distributions.

**Declaration of competing interest**

The authors declare that they have no known competing financial interests or personal relationships that could have appeared to influence the work reported in this paper.

**Acknowledgments**

This work is supported by the National Natural Science Foundation of China (No. 12272267, No. 52278411), the Shanghai Oriental Talents Top-Tier Program, and the Fundamental Research Funds for the Central Universities.

**CRediT authorship contribution statement**

Siqi Wang: Writing – review & editing, Writing – original draft, Visualization, Validation, Software, Resources, Project administration, Methodology, Investigation, Formal analysis, Conceptualization; Daobo Sun: Software, Project administration; Yizheng Wang: Investigation, Writing – review & editing; Yilong Zhang: Investigation, Writing – review & editing; Yabin Jin: Supervision; Xiaoying Zhuang: Supervision, Funding acquisition; Timon Rabczuk: Supervision, Writing – review & editing, Funding acquisition, Investigation

**Appendix A. Discussion on computational complexity**

The computational cost of GA-VINO mainly consists of two parts: the cost of point cloud sampling and grouping, and the cost of attention operations. In the sampling and grouping layers, the FPS and ball query operations have complexities of $O(N_g{}^2)$ and $O(N_sN_g)$, respectively. Compared with SDF-type geometry processing methods, this point cloud sampling and grouping procedure usually introduces lower additional computational overhead. Further discussion on this part can be found in Ref. [29].

For the attention cost, GA-VINO uses an encoder-decoder architecture to reduce the $O(N^2)$ complexity of conventional Transformer-based neural operators to a form associated with $N_s$. Specifically, in the geometry encoder, the cross-attention layers mainly operate between $N_s$ sampled tokens and $N_g$ boundary points, and this part has a complexity of $O(N_sN_g)$. The self-attention layers operate among $N_s$ latent tokens, with a complexity of $O(N_s{}^2)$. In the solution decoder, $N_q$ query points are first encoded into a query matrix of size $N_q \times d_e$, and then interact through cross-attention with the KEY and VALUE matrices of size $N_s \times d_e$ produced by the encoder. Therefore, the main complexity of the solution decoder is $O(N_qN_s)$. For the most basic case with homogeneous material and uniform loading, GA-VINO only uses the boundary geometry point cloud as the encoder input, and its computational complexity can be expressed as

$$O(N_sN_g + N_s{}^2 + N_qN_s). \tag{57}$$

In more complex Mindlin–Reissner plate problems, the numbers of points in the geometry, material, and load point clouds may be different. For the most complex case, where the material and load are represented by different point clouds, the total computational complexity can be expressed as

$$O[N_s(N_g + N_m + N_l + N_q) + 3N_s{}^2]. \tag{58}$$

This expression indicates that the complexity of GA-VINO mainly grows linearly with the number of input points and query points, while the quadratic complexity only appears among the relatively small number of latent tokens. Therefore, even in complex cases where geometry, material, and load point clouds are considered simultaneously, GA-VINO can still maintain good computational efficiency.

**Appendix B. Element-wise energy evaluation for Mindlin–Reissner plates**

In this work, the model is trained by minimizing a functional derived from the variational formulation of the governing PDE. Inspired by the finite element method, the variational formulation discretizes the computational domain and uses shape functions to evaluate derivatives and integrals analytically. This enables the model to be trained without labeled solution data. For clarity, we derive the expression of the loss function in this appendix. GA-VINO is designed to predict the transverse deflection and two rotations at a set of query points on the Mindlin–Reissner plate:

$$\mathbf{u}_{\Theta}^{(q)} = [w^{(q)}, \theta_x^{(q)}, \theta_y^{(q)}]^T. \tag{59}$$

The total potential energy is written as

$$\Pi = U_{\text{bend}} + U_{\text{shear}} - W_{\text{ext}}. \tag{60}$$

For the Mindlin–Reissner plate problem, the variational formulation involves only first-order derivatives. Therefore, three-node linear triangular elements are used for the discrete energy evaluation. Following an element-wise variational strategy, the plate domain is divided into $N_e$ triangular elements, and the total potential energy can be written as

$$\Pi = \sum_{e=1}^{N_e} (U_{\text{bend}}^e + U_{\text{shear}}^e) - W_{\text{ext}} \tag{61}$$

For a three-node triangular element, the nodal values of a scalar field $\phi$ are interpolated by linear shape functions as

$$\hat{\phi}^e(x,y) = \sum_{a=1}^{3} N_a(x,y)\phi_a^e = \mathbf{N}\phi^e, \tag{62}$$

where

$$\mathbf{N} = \begin{bmatrix} N_1 & N_2 & N_3 \end{bmatrix}, \ \phi^e = \begin{bmatrix} \phi_1^e & \phi_2^e & \phi_3^e \end{bmatrix}^T. \tag{63}$$

Since the gradients of the linear shape functions are constant within each element,

$$\nabla\hat{\phi}^e = \mathbf{B}_e \phi^e, \tag{64}$$

with

$$\mathbf{B}_e = \begin{bmatrix} N_{1,x} & N_{2,x} & N_{3,x} \\ N_{1,y} & N_{2,y} & N_{3,y} \end{bmatrix}. \tag{65}$$

The three Mindlin–Reissner unknowns are approximated as

$$\hat{w}^e = \mathbf{N}\mathbf{w}^e, \ \hat{\theta}_x^e = \mathbf{N}\theta_x^e, \ \hat{\theta}_y^e = \mathbf{N}\boldsymbol{\theta}_y^e. \tag{66}$$

The bending curvatures are obtained from the gradients of the rotation field as

$$\kappa_{xx}^e = \hat{\theta}_{x,x}^e, \ \kappa_{yy}^e = \hat{\theta}_{y,y}^e, \ \kappa_{xy}^e = \frac{1}{2}\left(\hat{\theta}_{x,y}^e + \hat{\theta}_{y,x}^e\right). \tag{67}$$

For an isotropic plate under the plane stress assumption, the element-wise material parameters are given by

$$\mu_e = \frac{E_e}{2(1+\nu_e)}, \ \lambda_e = \frac{E_e \nu_e}{(1-\nu_e^2)}. \tag{68}$$

For homogeneous plates, $E_e = E$ and $\nu_e = \nu$. For spatially varying material fields, $E_e$ and $\nu_e$ are taken as element-wise constants. The bending energy of element $e$ is then computed as

$$U_{\text{bend}}^e = \frac{1}{2}\frac{h^3}{12} A_e \left[ \lambda_e \left(\kappa_{xx}^e + \kappa_{yy}^e\right)^2 + 2\mu_e \left( (\kappa_{xx}^e)^2 + (\kappa_{yy}^e)^2 + 2(\kappa_{xy}^e)^2 \right) \right], \tag{69}$$

where $A_e$ is the area of the triangular element. The transverse shear strain is defined as

$$\gamma^e = \nabla\hat{w}^e - \hat{\theta}^e, \tag{70}$$

or equivalently,

$$\gamma_x^e = \hat{w}_{,x}^e - \hat{\theta}_x^e, \ \gamma_y^e = \hat{w}_{,y}^e - \hat{\theta}_y^e. \tag{71}$$

The corresponding shear energy is

$$U_{\text{shear}}^e = \frac{1}{2}\kappa_s \mu_e h \int_{\Omega_e} \gamma^e \cdot \gamma^e d\Omega, \tag{72}$$

since $w$ and the rotations are linearly interpolated, this integral can be evaluated analytically. Define

$$\mathbf{M}_e = \int_{\Omega_e} \mathbf{N}^T \mathbf{N} d\Omega = \frac{A_e}{12}\begin{bmatrix} 2 & 1 & 1 \\ 1 & 2 & 1 \\ 1 & 1 & 2 \end{bmatrix}, \tag{73}$$

and

$$\mathbf{m}_e = \int_{\Omega_e} \mathbf{N}^T d\Omega = \frac{A_e}{3}\begin{bmatrix}1\\1\\1\end{bmatrix}. \tag{74}$$

Then the shear energy can be written as

$$U_{\text{shear}}^e = \frac{1}{2}\kappa_s \mu_e h \left[ A_e (\nabla \hat{w}^e)^T (\nabla \hat{w}^e) - 2(\nabla \hat{w}^e)^T \begin{bmatrix} \mathbf{m}_e^T \boldsymbol{\theta}_x^e \\ \mathbf{m}_e^T \boldsymbol{\theta}_y^e \end{bmatrix} + (\boldsymbol{\theta}_x^e)^T \mathbf{M}_e \boldsymbol{\theta}_x^e + (\boldsymbol{\theta}_y^e)^T \mathbf{M}_e \boldsymbol{\theta}_y^e \right], \tag{75}$$

where $\nabla \hat{w}^e = \mathbf{B}_e \mathbf{w}^e$ is constant within the element. For a transverse pressure field $p(x,y)$, the external work acting on element $e$ is

$$W_{\text{ext}}^e = \int_{\Omega_e} p(x,y)\hat{w}(x,y)\mathrm{d}\Omega. \tag{76}$$

If the pressure is represented by nodal values

$$\mathbf{p}^e = \begin{bmatrix} p_1^e & p_2^e & p_3^e \end{bmatrix}^T, \hat{p}^e = \mathbf{N}\mathbf{p}^e, \tag{77}$$

the domain contribution can be evaluated analytically as

$$W_{\text{ext},\Omega}^e = (\mathbf{w}^e)^T \mathbf{M}_e \mathbf{p}^e. \tag{78}$$

For a prescribed transverse line load $q(s)$ acting on a subset $\Gamma_q$ of the plate boundary, the external work is evaluated through a boundary integral:

$$W_{\text{ext},\Gamma} = \int_{\Gamma_q} q(s)w(s)\mathrm{d}\Gamma = \sum_{b=1}^{N_{seg}} W_{\text{ext},\Gamma}^b, \tag{79}$$

where $N_{seg}$ denotes the number of two-node boundary segments on $\Gamma_q$. For each loaded boundary segment $b$, the transverse deflection and line load are interpolated using one-dimensional linear shape functions as

$$\hat{w}^b = \mathbf{N}_\Gamma^b \mathbf{w}^b,\ \hat{q}^b = \mathbf{N}_\Gamma^b \mathrm{q}^b \tag{80}$$

where $\mathbf{N}_\Gamma^b = \begin{bmatrix} N_1^b & N_2^b \end{bmatrix}$, $w^b = \begin{bmatrix} w_1^b & w_2^b \end{bmatrix}^T$ and $\mathbf{q}^b = \begin{bmatrix} q_1^b & q_2^b \end{bmatrix}^T$.

The external work associated with boundary segment $b$ can then be evaluated analytically as

$$W_{\text{ext},\Gamma}^b = (\mathbf{w}^b)^T \mathbf{M}_b \mathbf{q}^b, \tag{81}$$

where the boundary integration matrix is given by

$$\mathbf{M}_b = \int_{\Gamma_b} (\mathbf{N}_\Gamma^b)^T \mathbf{N}_\Gamma^b \mathrm{d}\Gamma = \frac{L_b}{6}\begin{bmatrix} 2 & 1 \\ 1 & 2 \end{bmatrix}, \tag{82}$$

and $L_b$ denotes the length of the boundary segment. Combining the domain and boundary contributions, the total external work can be expressed as

$$W_{\text{ext}} = \sum_{e=1}^{N_e} (\mathbf{w}^e)^T \mathbf{M}_e \mathbf{p}^e + \sum_{b=1}^{N_{seg}} (\mathbf{w}^b)^T \mathbf{M}_b \mathbf{q}^b. \tag{83}$$

The first term represents the external work of the distributed transverse pressure over the plate domain, whereas the second term corresponds to the prescribed transverse line load along the boundary. Either contribution vanishes when the corresponding loading type is absent. For the annular-sector plate considered in Section 4.4, no distributed pressure is applied, and only the boundary contribution is retained.

For a training set with $N_{\text{train}}$ input samples, the energy-based loss is defined as

$$\mathcal{L}(\boldsymbol{\Theta}) = \frac{1}{N_{\text{train}}} \sum_{i=1}^{N_{\text{train}}} \Pi^{(i)}, \tag{84}$$

where $\Pi^{(i)}$ denotes the discrete total potential energy of the $i$-th training sample, evaluated using Eqs. (61) and (83).

Thus, the nodal responses predicted by the network can be transformed into a differentiable discrete total potential energy through shape function interpolation and element-wise integration. This discrete total potential energy is used as the physics training objective to optimize the network parameters, without introducing finite element displacement or rotation labels as supervised terms. Therefore, the finite element solutions are used only for accuracy evaluation in the testing stage, while the training process is fully driven by the variational energy functional of the Mindlin–Reissner plate.

## Appendix C. Discussion on the effect of hyperparameters

GA-VINO involves several hyperparameters. To analyze the influence of different hyperparameter settings on model performance, we use the single-hole plate example with homogeneous material under uniform loading as a representative case. The effects of the number of attention layers, the point cloud sampling and grouping parameters $N_s$, $N_p$, and $r$ are discussed. To ensure a fair comparison and isolate the effect of each hyperparameter, early stopping is disabled in these experiments, and all models are trained for 500 epochs.

Table 8 summarizes the prediction accuracy and training efficiency of GA-VINO under different attention layer settings. The results indicate that encoder cross-attention plays an important role in model performance. Removing the encoder cross-attention block leads to a substantial increase in prediction errors, highlighting its importance in extracting geometric features from boundary point clouds and constructing effective latent representations. When the number of encoder cross-attention layers increases from 1 to 2 and further to 4, the prediction accuracy progressively improves, although the number of trainable parameters also increases. In comparison, encoder self-attention has a relatively smaller influence on prediction accuracy. Even without encoder self-attention, the model can maintain reasonable prediction accuracy, indicating that cross-attention already captures important geometric information in this single-hole plate example. Nevertheless, increasing the number of encoder self-attention layers further reduces the prediction errors, suggesting that information exchange among latent tokens contributes positively to model performance. For decoder cross-attention, increasing the number of layers from 4 to 8 also improves prediction accuracy, but leads to a noticeable increase in training time due to the large number of query points processed by the decoder. Overall, these results indicate that increasing the number of attention layers can improve prediction accuracy, but the additional benefits should be balanced against the increased model complexity and computational cost.

Table 8 Effect of the number of attention layers on prediction accuracy and training efficiency.

| Encoder/Decoder Cross-att. layers | Encoder Self-att. layers | $\lVert e_w \rVert_{L^2}$ (%) | $\lVert e_\theta \rVert_{L^2}$ (%) | $\lVert e_w \rVert_{H^1}$ (%) | $\lVert e \rVert_E$ (%) | Trainable parameters | Training time (s) |
|---|---|---|---|---|---|---|---|
| 0/4 | 2 | 4.26 | 8.59 | 8.80 | 16.94 | 677,162 | 2,964.33 |
| 1/4 | 2 | 1.98 | 4.93 | 4.34 | 10.98 | 827,398 | 3,016.84 |
| 2/4 | 0 | 2.74 | 6.16 | 6.30 | 15.17 | 677,930 | 3,035.94 |
| 2/4 | 2 | 1.43 | 3.31 | 3.75 | 9.63 | 977,834 | 3,019.01 |
| 4/4 | 2 | 1.26 | 2.85 | 3.07 | 7.40 | 1,278,506 | 3,025.27 |
| 2/4 | 4 | 1.22 | 2.62 | 2.92 | 7.23 | 1,277,738 | 3,039.61 |
| 1/8 | 2 | 1.40 | 3.18 | 3.51 | 8.56 | 1,027,946 | 4,076.90 |

Table 9 analyzes the influence of the sampling and grouping parameters $N_s$, $N_p$, and $r$ on model performance. Overall, increasing $N_s$ and $N_p$ leads to a longer training time, which is consistent with the discussion on computational complexity in Appendix A. When these parameters are varied within a reasonable range, the $L^2$ error, $H^1$ seminorm error, and energy norm error of GA-VINO change only moderately, and the training time does not fluctuate significantly. This indicates that the model is not overly sensitive to the hyperparameters related to point cloud sampling and grouping, and exhibits good stability.

Table 9 Influence of sampling and grouping parameters on the prediction performance and training efficiency of GA-VINO.

| $N_s$ | $N_p$ | $r$ | $\lVert e_w \rVert_{L^2}$ (%) | $\lVert e_\theta \rVert_{L^2}$ (%) | $\lVert e_w \rVert_{H^1}$ (%) | $\lVert e \rVert_E$ (%) | Memory (MB) | Training time (s) |
|---|---|---|---|---|---|---|---|---|
| 16 | 18 | 0.2 | 1.88 | 4.12 | 4.72 | 11.76 | 5,685.67 | 1,813.30 |
| 64 | 18 | 0.2 | 1.53 | 3.49 | 3.90 | 9.69 | 6,643.80 | 2,031.94 |
| 128 | 18 | 0.2 | 1.38 | 3.09 | 3.39 | 8.45 | 8,565.67 | 2,339.25 |
| 256 | 18 | 0.2 | 1.43 | 3.31 | 3.75 | 9.63 | 12,917.35 | 3,019.01 |
| 384 | 18 | 0.2 | 1.55 | 3.42 | 3.78 | 9.57 | 17,301.60 | 3,654.24 |
| 256 | 18 | 0.1 | 1.32 | 3.00 | 3.28 | 8.11 | 12,922.69 | 3,054.87 |
| 256 | 18 | 0.3 | 1.40 | 3.32 | 3.65 | 9.21 | 12,922.69 | 3,043.79 |
| 256 | 6 | 0.2 | 2.48 | 5.48 | 5.89 | 14.10 | 12,741.42 | 3,044.11 |
| 256 | 64 | 0.2 | 1.51 | 3.30 | 3.78 | 9.32 | 13,616.74 | 3,120.44 |

Specifically, $N_s$ controls the number of latent tokens used in the geometry encoder to represent the boundary geometry. When $N_s$ is too small, the sampled points may be insufficient to fully cover complex boundaries, thereby weakening the geometric representation capability of the model. When $N_s$ is further increased, the computational cost increases, but the prediction accuracy does not necessarily improve further. This suggests that a moderate value of $N_s$ is already sufficient to provide an effective geometric representation in the present example. The parameter $N_p$ determines the number of points contained in the local neighborhood of each sampled point. When $N_p$ varies within a relatively small range, its influence on the prediction results of GA-VINO is limited. However, when $N_p$ becomes too large, a decrease in prediction accuracy can be observed, indicating that an excessively large $N_p$ may introduce redundant points or overly broad local neighborhoods, thereby reducing the effectiveness of local geometric features. The search radius $r$ controls the spatial scale of local grouping. When $r$ is too small, the local neighborhood may fail to cover enough boundary points. When $r$ is too large, the local group may contain excessive nonlocal information, reducing the discriminability of local geometric features.

## Appendix D. FEM convergence and mesh-quality sensitivity study

To verify the reliability of the finite element reference solutions used in this work, mesh-convergence and mesh-quality sensitivity analyses are performed for a simply supported square Mindlin–Reissner plate with a Navier-type analytical solution. The FEM solution is obtained using the Durán–Liberman mixed Mindlin–Reissner plate formulation implemented in DOLFINx. This formulation belongs to the MITC-type family of shear-locking-free methods, in which the transverse shear strain is treated through mixed interpolation to improve the numerical stability of Mindlin–Reissner plate analysis for thin and moderately thick plates. Specifically, the transverse deflection $w$ is discretized using first-order continuous Lagrange elements, the rotation field is discretized using second-order continuous vector-valued Lagrange elements, and the auxiliary fields associated with the shear strain are discretized using suitable mixed finite element spaces. More details of this formulation can be found in Ref. [33]. After the computation, the finite element nodal solution is compared with the Navier-type analytical solution for error evaluation.

A linearly elastic Mindlin–Reissner plate defined on the square domain $\Omega \subset [0,5]^2$ is considered. The plate thickness is $h = 0.5$ mm, and the material parameters are $E = 1000$ MPa and $\nu = 0.3$. Simply supported boundary conditions are imposed on the four edges, which can be written as

$$\begin{aligned} &w = 0, \theta_y = 0, (\text{at x=0, x=5}), \\ &w = 0, \theta_{\text{x}} = 0, (\text{at y=0, y=5}). \end{aligned} \tag{85}$$

The plate is subjected to the following single-term sinusoidal transverse pressure load

$$p(x, y) = p_0 \sin \frac{(\pi x)}{a} \sin \frac{(\pi y)}{b}. \tag{86}$$

Here, $a = b = 5$ mm denote the side lengths of the square plate. Since the load contains only one sinusoidal mode, the following Navier-type analytical expressions for the displacement and rotations can be expressed as [41]

$$\begin{aligned} &w(x, y) = \vartheta \sin \frac{(\pi x)}{a} \sin \frac{(\pi y)}{b}, \\ &\theta_x(x, y) = \Theta_x \cos \frac{(\pi x)}{a} \sin \frac{(\pi y)}{b}, \\ &\theta_y(x, y) = \Theta_y \sin \frac{(\pi x)}{a} \cos \frac{(\pi y)}{b}. \end{aligned} \tag{87}$$

Here $\vartheta$, $\Theta_x$, and $\Theta_y$ are the amplitude coefficients of the analytical solution. This form automatically satisfies the simply supported boundary conditions of the square plate.

For the mesh convergence study, the square plate is discretized using triangular meshes with progressively increasing element densities. To examine the reliability of the FEM reference solution for different plate thicknesses, two thicknesses, $h = 0.5$ mm and $h = 0.005$ mm, are considered. The corresponding load amplitudes are set to $p_0 = 0.8$ MPa and $p_0 = 8 \times 10^{-7}$ MPa, respectively, so that the deflection magnitudes remain comparable between the two thickness cases. The FEM solutions obtained at different mesh resolutions are compared with the corresponding Navier-type analytical solutions. Fig. 20 (a) and (b) show the relative $L^2$ errors of the deflection and rotation fields, the relative $H^1$ seminorm error of the deflection field, and the energy-norm error as the mesh is progressively refined.

The sensitivity of the FEM solution to triangular-element quality is further examined for the $h = 0.5$ mm case. Five meshes with different element-quality distributions are generated while the total number of triangular elements is fixed at 20,000, so that the influence of element distortion can be examined independently. Element quality is characterized using the element-level metric $q_e$ and the mesh-level indicator $q_{5\%}$ defined in Section 4.9.2. Fig. 20 (c) presents the corresponding errors as a function of $q_{5\%}$, while Fig. 20 (d) shows representative meshes at different quality levels.

Fig. 20 (a) and (b) show clear convergence of the FEM solutions with increasing mesh resolution for both plate thicknesses. Even for the coarsest meshes containing only 800 triangular elements, the $L^2$ and $H^1$ seminorm errors remain below 1% and decrease further as the meshes are refined. The energy-norm errors are slightly larger than the $L^2$ and $H^1$ errors but remain within an acceptable range. These results demonstrate that the adopted FEM formulation provides high solution accuracy over the range of plate thicknesses considered.

The mesh-quality sensitivity results in Fig. 20 (c) further show that the FEM solutions are only weakly affected by variations in triangular-element quality when the mesh resolution is kept fixed. The $L^2$ and $H^1$ seminorm errors remain small over the entire range of $q_{5\%}$ considered. The energy-norm error shows a somewhat stronger dependence on element quality, but remains below 1% for all five meshes. Therefore, within the range investigated in this study, element distortion does not significantly affect the FEM solution.

Overall, the mesh-convergence and mesh-quality sensitivity studies demonstrate that the FEM formulation adopted in this work provides sufficiently accurate and numerically robust reference solutions for evaluating the prediction accuracy of GA-VINO.

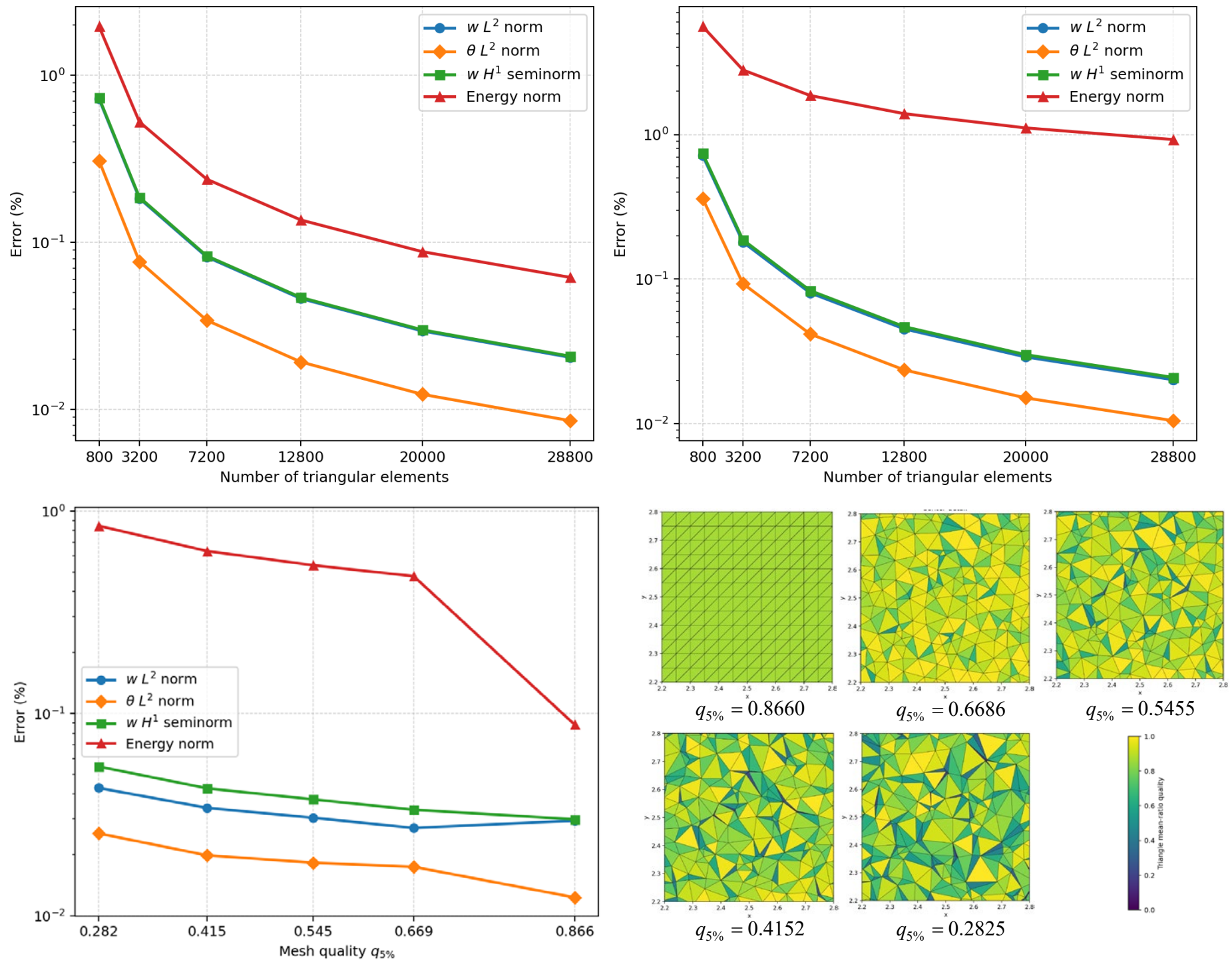

Fig. 20. FEM convergence and mesh-quality sensitivity analyses. (a) FEM convergence for $h = 0.5$ mm ; (b) FEM convergence for $h = 0.005$ mm ; (c) influence of triangular-element quality for $h = 0.5$ mm with 20,000 elements in each mesh; (d) representative meshes with different quality levels used in the sensitivity analysis.

## Appendix E. Resolution invariance analysis

One important advantage of neural operators is resolution invariance [13][14], meaning that the model learns a mapping between input and output functions rather than a point-to-point mapping on a fixed grid or a fixed set of discretization points. Therefore, a model can be trained at one resolution and evaluated at different query point densities during testing. Here, the resolution invariance of GA-VINO is investigated using the case in Section 4.1.1 as the benchmark. During training, the query points also serve as the nodes of the triangular discretization used for evaluating the variational energy, and the corresponding query-point density is denoted as 50%. In the present resolution-invariance study, the trained model parameters and the training-stage triangulation remain unchanged. Only the query-point set supplied to the decoder during testing is varied. Specifically, the same trained model is directly evaluated on 100 test samples using ten query-point densities ranging from 10% to 100%.

Fig. 21 shows the predicted transverse deflection and rotation fields of GA-VINO under different query point densities. The results indicate that GA-VINO exhibits strong resolution invariance as the query point density varies. However, when the query point density is too low, insufficient spatial sampling may lead to the loss of some high-frequency details and local features.

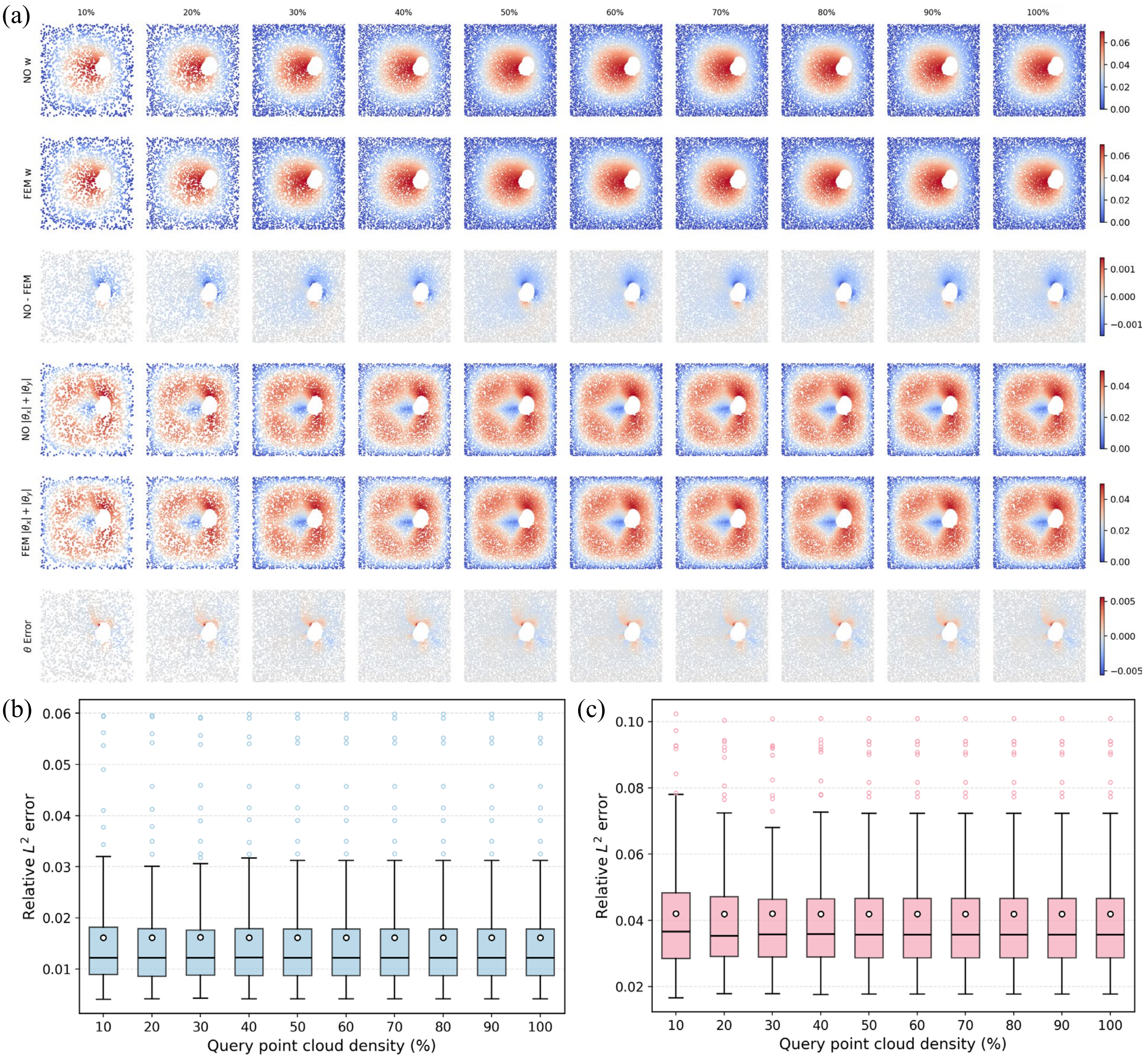

Fig. 21. Resolution invariance evaluation of GA-VINO. (a) Predicted transverse deflection and rotation fields at different query point densities, with the density increasing from 10% to 100% at intervals of 10%. Results of (b) $\|e_w\|_{L^2}$ (%) and (c) $\|e_\theta\|_{L^2}$ (%) under different query point densities.

## Appendix F. Supplementary code

The code for this work is available at https://github.com/siqi-wang915/GA-VINO.